\documentclass[runningheads]{llncs}

\usepackage{eccv}

\usepackage{eccvabbrv}

\usepackage{graphicx}
\usepackage{booktabs}

\usepackage[accsupp]{axessibility}  % Improves PDF readability for those with disabilities.

\usepackage[pagebackref,breaklinks,colorlinks,citecolor=eccvblue]{hyperref}
\usepackage{orcidlink}

\usepackage{caption,subcaption}
\usepackage{tikz}
\usepackage{comment}
\usepackage{amsmath,amssymb} % define this before the line numbering.
\usepackage{color}
\usepackage{xcolor}
\usepackage{arydshln}
\usepackage{amsmath}
\usepackage{amssymb}
\usepackage{multirow}
\usepackage{makecell}
\usepackage{tablefootnote}
\usepackage{pifont}
\usepackage{xr}
\usepackage{overpic, textpos}
\usepackage{algorithmicx}
\usepackage{colortbl}
\usepackage{tabulary}
\usepackage{algpseudocode}
\usepackage{listings}
\usepackage{xspace}
\usepackage{wrapfig}
\usepackage{algorithm}
\usepackage{adjustbox}
\usepackage{wrapfig}
\usepackage{subfiles}

\newlength\secmargin
\newlength\subsecmargin
\newlength\subsubsecmargin
\newlength\paramargin
\newlength\abovetabcapmargin
\newlength\belowtabcapmargin
\newlength\abovefigcapmargin
\newlength\belowfigcapmargin
\newcolumntype{L}[1]{>{\raggedright\let\newline\\\arraybackslash\hspace{0pt}}m{#1}}
\newcolumntype{C}[1]{>{\centering\arraybackslash}m{#1}}
\newcolumntype{R}[1]{>{\raggedleft\let\newline\\\arraybackslash\hspace{0pt}}m{#1}}

\newcommand{\ournet}{SPDP-Net}
\newcommand{\tr}{DPATr}
\newcommand{\cmark}{\ding{51}}%
\newcommand{\xmark}{\ding{55}}%
\begin{document}

% \title{Spatio-Temporal Proximity Representations for Panoramic Human Activity Recognition?} 
%\title{Breaking Hierarchical Boundaries: Spatio-Temporal Proximity-Aware Dual-Path Model for Panoramic Human Activity Recognition}
\title{Spatio-Temporal Proximity-Aware Dual-Path Model for Panoramic Activity Recognition} 

%\title{Spatio-Temporal Proximity Representations and Multigranular Dual-Path for Panoramic Human Activity Recognition}

\titlerunning{SPDP-Net}

\author{Sumin Lee %\inst{1,2}
\and
Yooseung Wang %\inst{1} 
\and
Sangmin Woo %\inst{1}
\and
Changick Kim %\inst{1}
}

\authorrunning{S. Lee~\textit{et al}.}

\institute{Korea Advanced Institute of Science and Technology (KAIST), Daejeon, Korea
\email{\{suminlee94, yswang, smwoo95, changick\}@kaist.ac.kr}}

\maketitle

\begin{abstract}
Panoramic Activity Recognition (PAR) seeks to identify diverse human activities across different scales, from individual actions to social group and global activities in crowded panoramic scenes.
PAR presents two major challenges: 1) recognizing the nuanced interactions among numerous individuals and 2) understanding multi-granular human activities.
To address these, we propose Social Proximity-aware Dual-Path Network (SPDP-Net) based on two key design principles.
First, while previous works often focus on spatial distance among individuals within an image, we argue to consider the spatio-temporal proximity.
It is crucial for individual relation encoding to correctly understand social dynamics.
Secondly, deviating from existing hierarchical approaches (individual-to-social-to-global activity), we introduce a dual-path architecture for multi-granular activity recognition.
This architecture comprises individual-to-global and individual-to-social paths, mutually reinforcing each other's task with global-local context through multiple layers.
Through extensive experiments, we validate the effectiveness of the spatio-temporal proximity among individuals and the dual-path architecture in PAR.
Furthermore, SPDP-Net achieves new state-of-the-art performance with 46.5\% of overall F1 score on JRDB-PAR dataset.


\end{abstract}

\section{Introduction}
Understanding human activity in videos is a pivotal task in computer vision.
It finds diverse real-world applications across various domains, including video sports analysis~\cite{sports_analysis1,sports_analysis2,sports_analysis3}, surveillance~\cite{vid_surveil1,vid_surveil2}, and social scene analysis~\cite{social_analysis,social_analysis2,social_analysis3}.
Most primary research has focused on recognizing behaviors at different granularity levels, such as Human Activity Recognition (HAR)~\cite{ar_survey,multi4,modality_mixer,2stream:tsn}, Group Activity Recognition (GAR)~\cite{gar_survey,gar1,wgar1,wgar2}, and Panoramic Activity Recognition (PAR)~\cite{par_eccv,mup}. 
HAR systems concentrate on identifying the actions of a single individual in well-defined settings.
The task of GAR extends the scope of HAR by focusing on understanding interactions and collective activities performed by multiple individuals in a shared space.
Building on the concept of GAR, PAR tackles an even more comprehensive problem to understand human behavior in complex environments.
PAR aims to recognize social group activities as well as individual and collective activities in a panoramic scene.
The objectives of PAR involve recognizing multi-granular activities through the three tasks: ({\romannumeral 1}) individual action recognition, ({\romannumeral 2}) social group activity recognition coupled with social group detection, and ({\romannumeral 3}) global (panoramic) activity recognition. 
 
\begin{figure}[t]
    \centering
    \includegraphics[width=\textwidth]{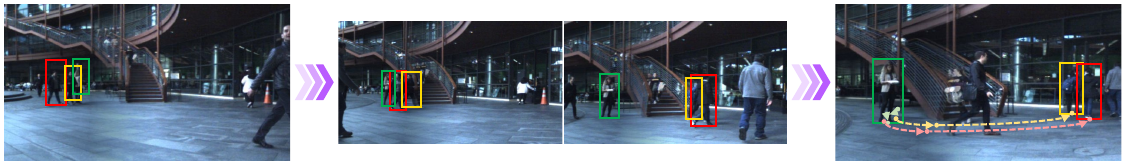}
    \vspace{-0.7cm}
    % \vspace{\abovefigcapmargin}
    \caption{
    Importance of the spatio-temporal proximity for understanding social group dynamics.
    To distinguish between social groups, it is crucial to leverage positional relationships among individuals not just in space but also over time. 
    Consider an initial scene where individuals marked with \textcolor{red}{red}, \textcolor{yellow!40!orange}{yellow}, and \textcolor{green}{green} bounding boxes are close to each other, giving the impression that they belong to the same social group.
    However, as time goes on, it becomes evident that only the individuals in the \textcolor{red}{red} and \textcolor{yellow!40!orange}{yellow} boxes move together, indicating shared social group membership, while the person in the \textcolor{green}{green} box does not.
    % Thus, it is essential to leverage spatio-temporal proximity for precise social group recognition.
    % By leveraging spatio-temporal positional relationships, the proposed \ournet\ correctly detects social groups and predicts social group activities. 
    }
    \label{fig:teasure}
    \vspace{\belowfigcapmargin}
\end{figure}

PAR presents a unique set of challenges.
Firstly, due to the intricate nature of panoramic scenes, those scenes often contain numerous individuals engaged in diverse activities and interactions within a wide field of view.
For this reason, previous works~\cite{par_eccv,mup} utilized spatial proximity in a single frame for determining social relations between individuals.
% However, we argue that the distances between individuals in a still image are insufficient for defining social dynamics, since the distances vary over time.
However, we argue that relying solely on spatial proximity is insufficient; it is imperative to incorporate spatio-temporal proximity in PAR.
% For this reason, previous works~\cite{par_eccv, mup} utilized spatial proximity in a single frame for group detection, which influences the determination of group dynamics and interaction.
% the spatial proximity among individuals is a critical factor for accurately identifying actions in these scenes.
% Moreover, the temporal context plays a key role in PAR, since it involves discerning multiple activities in a dynamic environment.
For example, as depicted in Fig.~\ref{fig:teasure}, the three individuals in the first frame seem to belong to the same social group due to their closeness.
A few frames later, the two in the yellow and red bounding boxes are moving together, whereas the individual in the green box is not.
We can see that only two in the yellow and red boxes share social membership, while the one in the green box does not.
% For example, in Fig.~\ref{fig:teasure}, the three individuals appear very close in the first frame, giving the impression that they belong to the same group. 
% However, a few frames later, it becomes evident that the two in the yellow and red boxes are moving together, while the individual in the green box is not, indicating that they do not belong to the same social group.
Therefore, it is essential to consider spatial position variations over time to accurately assess the social proximity for precise understanding of social dynamics.
Another challenge lies in comprehending the multi-granular activities ranging from simple individual actions to complex social group and global activities.
% These activities range from simple individual actions to complex social group and global activities.
Previous works~\cite{par_eccv,mup} hierarchically model three granular activities: `from individuals to groups', and `from groups to global'.
From these studies, we observe that both global and social group activities require individual contextual information, while they also mutually influence each other.
% to adeptly handle diverse social group sizes and complex dependencies among multi-granular activities, it is crucial to incorporate global-local context from both individual and global activities. 
We empirically validate this observation (see Sec.~\ref{sec:dptr}).

%======================================================================

% Previous works for PAR~\cite{par_eccv,mup} conceptualized the PAR problem involving three tasks as a hierarchical structure.
% They merge extracted individual features to social group features and social group features to global features, sequentially.
% In encoding social group features, spatial distance and affinity matrices are used to represent social relation among individuals.
% The spatial distance is measured by normalized euclidean distance in a single frame.
% But, in actual scene, determining physical identities solely based on the straight-line distance in a single frame is demanding.
% \textcolor{blue}{However, on analysis of their experimental results~\cite{mup}, it has been demonstrated that individual information is more effective in predicting global activities compared to social group information. 
% Also,to accurately predict social group clustering and the activities within those clusters, a fundamental understanding of the local-global context is essential.}
% For multi-granular activity recognition, it is essential to have a thorough comprehensive understanding of the atomic unit, individual's action information.
% Both intra- and inter-individual relations prove advantageous in discerning higher granular activities. 
% Especially within panoramic scenes, characterized by a wide view and a multitude of individuals, spatial proximity across time emerges as a significant factor in shaping individual interactions~\cite{physical1,physical2}.

To address these problems, we propose a novel network, called Social Proximity-aware Dual-Path Network (\ournet).
Our \ournet\ consists of two stages: 1) individual relation encoding and 2) multi-granular activity recognition.
In the individual relation encoding stage, spatio-temporal positional relationships among individuals are considered to refine their feature representations, enabling the precise measurement of social proximity.
This allows \ournet\ to capture accurate social dynamics within a panoramic scene.
Specifically, a spatio-temporal self-attention mechanism is employed on the features of each individual to emphasize crucial visual clues such as related objects and body posture.
% To further enhance this process and consider the spatio-temporal position of each person within a panoramic scene, we introduce a panoramic positional embedding,  applied to each self-attention layer.
To further enhance this process and account for overall contextual information within a panoramic scene, we incorporate a panoramic positional embedding that encapsulates the spatio-temporal positional information of each individual across the entire scene.
Additionally, we evaluate the social interactions among individuals for social group detection by measuring the feature similarity and the social proximity relation.
To accurately measure social proximity among individuals, we extend the concept of GIoU to the temporal axis, named Temporal Generalized IoU (GIoU).
For the multi-granular activity recognition, we introduce Dual-Path Activity Transformer (\tr).
Different from previous works that model multi-granular activities hierarchically~\cite{par_eccv,mup}, each layer of \tr\ consists of two paths, individual-to-global and individual-to-social paths.
The individual-to-global path concurrently encodes individual and global activities, exploiting the global-local context of the given video.
Then, the individual-to-social path explores social group activity information from globally attended individual features.
Through multiple layers, \tr\ mutually reinforces contextual understandings of multi-spatial activities, creating synergistic effects that enhance final predictions.
% This allows \tr\ to harness the harmonious effects of both global and local texts, improving performances through  synergistic interactions on multi-spatial activities.
% \ournet\ uniquely integrates insights into both individual and collective activities to effectively leverage both global and local context.
% This enables \tr\ to improve the recognition not only of global activities but also of social group activities.
% The cornerstone of our approach lies in the innovative integration of spatio-temporal positional cues with global-local contextual understanding.
% This integration is pivotal for capturing the intricacies of human interactions and activities in a panoramic scene.

To evaluate \ournet, we conduct extensive experiments on JRDB-PAR~\cite{par_eccv}.
Throughout the performance improvements in both human activity recognition and social group detection, we demonstrate the effectiveness of utilizing spatio-temporal proximity to model social dynamics in a wide crowded scene.
% The results show the effectiveness of utilizing spatio-temporal proximity to model human activity relationships in crowded scenes.
Moreover, we show that our \tr\ mechanism cooperatively strengthens the tasks of PAR by adeptly capturing the global-local contextual information.
% he semantic relations among three tasks (\ie, individual action, social group activity, and global activity recognition).
Notably, the proposed \ournet\ significantly outperforms the state-of-the-art methods by a large margin, achieving 46.4\% of an overall F1 score for activity recognition and 56.4\% of IoU@0.5 for social group detection.

% The proposed STPC-Net excavates not only spatial-temporal but also physical (spatial proximity) relations of inter- and intra-individual to address three different granularity activities.
% First, through self-attention in individual visual features, STPC-Net leverages spatio-temporal information of atomic individual action.
% Subsequently, not only similarity of those individual features but also physical relation (\ie, spatial proximity) are measured for social group detection.
% For calculating physical relation, we use temporally averaged GIoU of given individual bounding boxes.
% GIoU can represent the extent of overlap and the distance between two bounding boxes.
% To classify individual actions, social group activities, and global activity, we introduce Multi-Granular Activity Transformer (MGATr).
% \textcolor{red}{[explain of MGATr]}

Our main contributions are summarized as follows:
\renewcommand\labelitemi{\tiny$\bullet$}
\begin{itemize}
% \item We investigate how to take interactions between individuals based on changes in their position for PAR problem.
% \item We analyze the semantic relationships among three multi-granular activities (\ie individual action, social group activity, and global activity) in PAR.
\item In this paper, we propose a novel network, named \ournet. For comprehensively understanding social dynamics in a wide crowded scene, \ournet\ leverages spatio-temporal proximity among individuals by utilizing a panoramic positional embedding and TGIoU.
\item Moreover, \tr\ in \ournet\ mutually reinforces collaboration among multi-granular activity tasks by utilizing a global-local context with a dual-path architecture.
% The proposed \ournet\ captures the spatio-temporal proximity among individuals and global-local contextual information to model human activities from various granularities in a crowded panoramic scene.
\item Throughout extensive experiments, we demonstrate the effectiveness of the proposed method. Furthermore, \ournet\ significantly outperforms the state-of-the-art method, achieving 46.5\% in the overall F1 score for activity recognition and 56.4\% in IoU@0.5 for social group detection.
\end{itemize}
\section{Related Work}
% \vspace{\subsubsecmargin}
\subsubsection{Human Action Recognition.}
Comprehending human actions is a fundamental task in the field of video understanding.
The task of action recognition aims to recognize the action classes performed by a single actor in a given trimmed video.
Earlier works focus on utilizing optical flow in two-stream architecture~\cite{2stream:tsn,2stream1}.
Thanks to the recently improved capacity of deep learning architecture, 3D convolution networks~\cite{3d_conv:1i3d,3d_conv:nonlocal,3d_conv:slowfast,3d_conv:stconv} or transformer-based networks~\cite{vivit,vidswin} are proposed, which lead to significant performance improvements.
ViViT~\cite{vivit} extracts visual appearance and pose features from tubelet embeddings of an input video with transformer layers~\cite{transformer}.
Following these works, we also focus on understanding human.
However, our focus lies in analyzing multi-granular human activity in videos with multiple individuals, necessitating consideration of social interactions among individuals.
% Building upon this approach, we utilize individual self-attention to delve into the actions of each individual within a panoramic scene.
% To mitigate positional relationships of cropped individual features in an entire wide scene, we introduce the panoramic positional embeddings, specifically designed to retain the positional integrity of individuals within a panoramic scene.
% This task can be regarded as a video classification problem.
% Existing methods for action recognition can be categorized in three types: the appearance-based~\cite{3d_temp:slowfast,3d_temp:i3d,2stream:tsn,vivit,vidswin}, skeleton-based~\cite{skel1,skel2,skel3}, and multi-modal methods~\cite{modality_mixer,multi1,multi3,multi4}.
% The video datasets used in this task, such as UCF~\cite{ucf}, ActivityNet~\cite{activitynet}, NTU-RGBD~\cite{ntu60,ntu120}, are usually collected from the actions performed by one or two actors.

\vspace{\subsubsecmargin}\subsubsection{Group Activity Recognition with Human Interaction.}
% human interaction modeling을 하는 group activity recognition을 위주로 작성
% 어떤 방식으로 modeling 했는지 위주로 서술
Group Activity Recognition (GAR) aims to identify the activity performed by a collective of individuals within a given video.
Unlike recognizing actions of single individual, GAR requires exploring the aggregation of these actions to reveal the collective behavior of the entire group.
Many studies incorporate the individual action labels as an auxiliary supervision~\cite{at,sa-gat} to model the relationships among multiple actors for richer representations of group dynamics~\cite{rltn_model1,rltn_model2,dynamic,wgar1,wgar2}.
Graph-based network~\cite{din, arg} and Transformer-based network~\cite{groupformer,wgar2,at} are two common approaches for constructing potential interactions among individuals.
% Wu \etal~\cite{arg} proposed ARG method, which learn the appearance and positional relationships between actors.
% DIN~\cite{din} incorporates a person-specific interaction graph for exploiting spatio-temporal interactions between individuals.
Actor-transformer~\cite{at} selectively exploits group activity information from actor-specific static and dynamic representations.
Li \etal~\cite{groupformer} introduce Groupformer, which is a spatio-temporal transformer with clustered attention to jointly augment the individual and group representations. 
These studies show the effectiveness of transformer architecture in capturing the relationships between individual actions and collective activities.
% In our work, we aims to multi-spatial activities including social group activities as well as individual actions and collective activities.
In contrast to these works that focus on identifying collective activities, we propose Dual Path Activity Transformer (DPATr) for recognizing multi-granular activities including individual actions, social group activities, and global activities. 
% Additionally, to recognize social group activities, MGATr incorporates intra-group relationships derived from global-local context.
% These models have been demonstrated on sport datasets (\eg, NBA~\cite{nba} and Volleyball~\cite{volleyball}) and daily-life datasets (\eg, CAD~\cite{cad}).
% These datasets typically contain video clips labeled with a single group activity label.
% % This GAR task also can be viewed as a video classification problem.
% However, real-world scenarios generally comprise multiple groups of people with different social activities.

\vspace{\subsubsecmargin}\subsubsection{Social Activity and Group Detection.}
The goal of social activity and group detection is to divide crowds of people into distinct social groups as well as to understand the social activities of these groups.
While initial methods~\cite{groupdet1, groupdet3} focus solely on detecting the social group, recent works have expanded to recognize the social activities with new datasets~\cite{sa-gat,jrdb,jrmot, jrdb_act}. 
Some works~\cite{jrdb_act, sa-gat} model relationships of individuals with adjacency matrices in a panoramic scene.
In these works, individual visual features cropped from a panoramic image are used for graph clustering~\cite{spectral} and aggregated to social features for detecting social group activities.
% To detect social group activities, those works use graph clustering~\cite{spectral} and aggregate individual visual features cropped from a panoramic image.
Contrary to these approaches solely relying on cropped visual features, we argue that the proximity among individuals over both space and time is vital for discerning group dynamics and individual interactions.
Considering this, our \ournet\ adeptly utilizes spatio-temporal positional relationships among individuals to capture the positional context of cropped individual regions within an entire panoramic scene.
% Although spectral clustering can capture complex cluster structures, it is sensitive to parameters such as the number of clusters or the similarity measure used.
% To improve the performance of group detection, Ehsanpour \etal~\cite{jrdb_act} used an eigenvalue-based loss function for the similarity matrix and loss partitioning approach.
% the social grouping performance and to reduce the discrepancy between train and inference compared to [13], we propose to incorporate an eigenvalue-based loss function [10]
% on the similarity matrix extracted from the visual features
% and geometrical relations between the detected bounding
% boxes

\vspace{\subsubsecmargin}\subsubsection{Panoramic Activity Recognition.}
Recently, Panoramic Activity Recognition (PAR) has gained attention in the field of video understanding.
The PAR task involves recognizing what individuals are doing (individual actions), how groups of people interact (social group activities), and what is happening in an overall panoramic video (global activity).
Han \etal~\cite{par_eccv} introduced a hierarchical graph network based on GCN.
This allows for nodes at various levels to correspond to individual, group, and global activities, sequentially.
Following this, Cao \etal~\cite{mup} proposed Multi-granularity Unified Perception (MUP) framework, which is hierarchically applied to facilitate aggregation and interpretation across the spectrum of granularity.
% captures both intra-granularity (within a single level of activity) and cross-granularity (across different levels of activity) semantics.
% MUP operates hierarchically, facilitating the aggregation and interpretation across the spectrum of granularity.%, `from parts to individuals', `from individuals to groups', and `from groups to global'.
Those works utilize both individual and group-level features for global activity recognition, while a reverse stream that uses global features for social group-level and individual-level is employed in~\cite{par_eccv}.
From these, we can see the importance of individual contextual information for both global and social group activities and their mutual influence on each other.
To promote active interactions among multi-granular activities, \tr\ consists of individual-to-global and individual-to-social paths.
This architecture explores global-local context of a panoramic scene by using of individual context for social group and global activity recognition.
\section{Proposed Method}
 
%======================================================================
\begin{figure}[t]
    \centering
    \includegraphics[width=\textwidth]{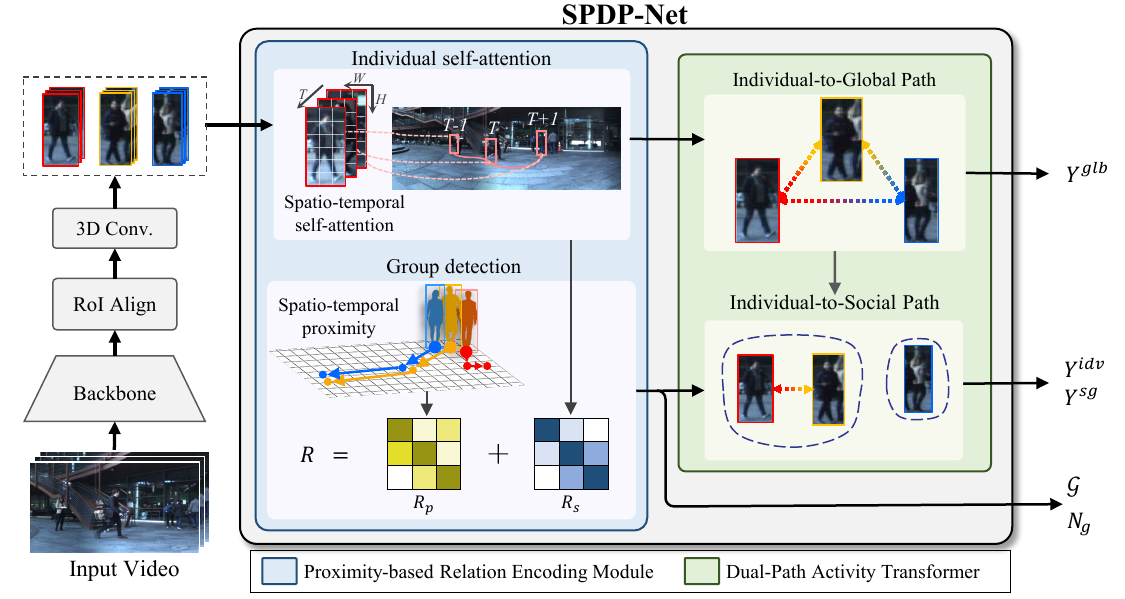}
    \vspace{-0.5cm}
    % \vspace{\abovefigcapmargin}
    \caption{Overview of the proposed \ournet. \ournet\ consists of two stages: 1) proximity-based relation encoding and 2) multi-granular activity recognition. $T_0$ indicates the center frame of a given video.
    }
    \label{fig:overall}
    \vspace{\belowfigcapmargin}
\end{figure}
%======================================================================

\subsubsection{Overview.}
\ournet\ aims to exploit spatio-temporal proximity representations of individuals with global-local contexts in a panoramic scene for PAR task.
The overview of \ournet\ is illustrated in Fig.~\ref{fig:overall}.
Given a panoramic video of length $T$, a 2D CNN backbone extracts frame-wise features.
To crop the appearance representations of individuals, we apply RoIAlign~\cite{roi_align} to each frame feature, followed by a 3D convolution for dimension reduction.
We denote the resulting feature for the entire spatio-temporal individual features as $F^{idv} \in \mathbb{R}^{N_i \times T \times d \times h \times w}$.
Here, $N_i$ denotes the number of individuals, $d$ is the hidden dimension, and $h$ and $w$ indicate the height and width of the cropped features, respectively.

\ournet\ consists of two stages: 1) proximity-based relation encoding and 2) multi-granular activity recognition.
% individual relation encoding이 하는 역할 설명
In the proximity-based relation encoding stage, spatio-temporal positional relationships are leveraged to capture the positional context of cropped individual regions in a panoramic scene.
Additionally, we measure the feature similarity and spatio-temporal proximity among individuals to understand social relationships among individuals.
% In the proximity-based relation encoding stage, spatio-temporal positional relationships among individuals are leveraged to comprehensively understand individual actions and social dynamics.
% This is achieved by processing $F^{idv}$ through a self-attention with spatio-temporal proximity information, resulting in $\bar{F}^{idv}$.
% Further explanation is described in Sec.~\ref{sec:3.2}.
For the multi-granular activity recognition, we introduce Dual-Path Activity Transformer (\tr), which models dependencies across three levels of granular activities with the individual-to-global and individual-to-social paths.
From activity features computed in \tr, dedicated classifiers for each granularity level predict activity scores:  $Y^{idv} \in \mathbb{R}^{N_i \times C_{idv}}$ for individual activities, $Y^{sg} \in \mathbb{R}^{n_g \times C_{sg}}$, for social group activities, and $y^{glb}\in \mathbb{R}^{C_{glb}}$ for global activity, where $C_{i}$ denotes the number of activity classes at each level.
% MGATR generates activity features for individuals ($\tilde{F}^{idv}\in \mathbb{R}^{N_i \times d}$), social groups ($F^{sg}\in \mathbb{R}^{N_g \times d}$), and global ($f^{glb}\in \mathbb{R}^{d}$).
% each granularity by taking $\bar{F}^{idv}$ and $\mathcal{G}$ as inputs.
% where , , and  respectively represent individual action, social group activity, and global activity features.
% Finally, dedicated classifiers for each granularity level predict activity scores:  $Y^{idv} \in \mathbb{R}^{N_i \times C_{idv}}$ for individual activities, $Y^{sg} \in \mathbb{R}^{n_g \times C_{sg}}$, for social group activities, and $y^{glb}\in \mathbb{R}^{C_{glb}}$ for global activity, where $C_{i}$ denotes the number of activity classes at each level.
% The detail explanations of MGATR are described in Sec.~\ref{sec:3.3}

\subsection{Proximity-based Relation Encoding}
\label{sec:3.2}
\begin{figure}[t]
\centering
     \begin{subfigure}{0.49\textwidth}
         \centering
         \includegraphics[width=\textwidth]{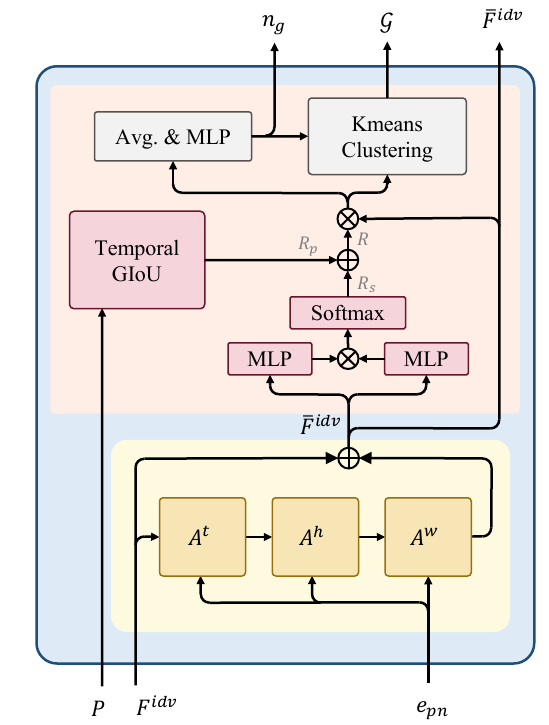}
         \caption{}
         \label{fig:idv_relation}
     \end{subfigure}
     \hfill
     \centering
     \begin{subfigure}{0.49\textwidth}
         \centering
         \includegraphics[width=\textwidth]{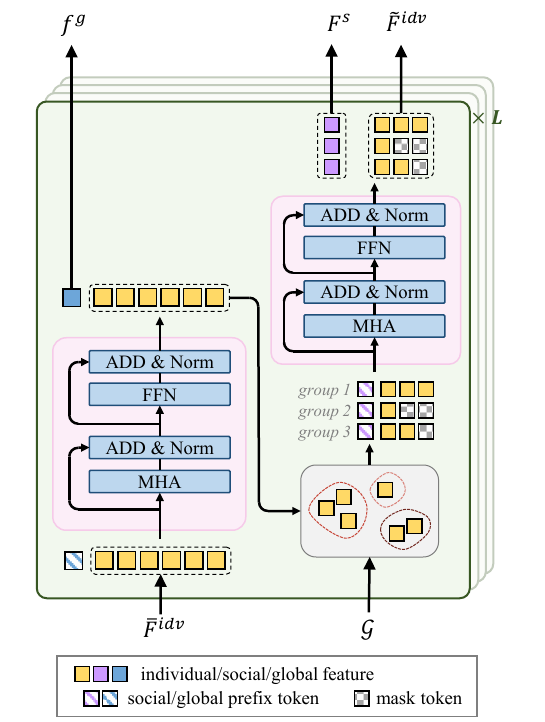}
         \caption{}
         \label{fig:mgar}
     \end{subfigure}
\vspace{-0.3cm}
% \vspace{\abovefigcapmargin}
\caption{Detailed architecture of two stages in \ournet. (a) Proximity-based relation encoding and (b) multi-granular activity recognition (\ie, \tr).}
\vspace{\belowfigcapmargin}
\label{fig}
\end{figure}

In this stage, the goal is to extract rich action information by considering the spatio-temporal proximity among individuals and to accurately perform social group detection.
The detailed mechanism of this stage is described in Fig.~\ref{fig:idv_relation}.
Firstly, we apply self-attention to the spatio-temporal individual feature $F^{idv}$.
The initial step is to apply spatio-temporal self-attention to each individual feature.
Motivated by~\cite{axial_attn}, we employ a Multi-Head Self-Attention (MHSA) along the temporal, height, and width dimensions in sequence.
A key challenge here is preserving the positional context of individuals within a wide panoramic scene.
However, conventional positional embeddings are inadequate in this aspect, since they only encode positional information within cropped regions.
% given that conventional positional embeddings tailored for cropped regions are inadequate in this aspect.
To address these, we incorporate the panoramic positional embedding, denoted as $e_{pn}$, specifically designed to retain the positional integrity of individuals within the panoramic scene.
We derive $e_{pn}$ by extracting individual regions from the sinusoidal positional embedding of the entire scene.
The self-attended features of individuals $\bar{F}^{idv}$ are calculated as follows:
\begin{align}
    \bar{F}^{idv} = A^w \left ( A^h \left ( A^t \left ( F^{idv}, e_{pn} \right ), e_{pn} \right ), e_{pn} \right ) + F^{idv},
\end{align}
where $A^t, A^h$, and $A^w$ indicate MHSA across the temporal, height, and width axes, respectively.

%======================================================================
%======================================================================

To detect social groups within the scene, our approach takes into account both the individual-specific features $\bar{F}^{idv}$ and the bounding box coordinates $P \in \mathbb{R}^{N_i \times T \times 4}$.
In practice, we measure the social connections between individuals by constructing a social relation matrix $R$ based on the visual similarities, denoted as $R_s$, and the spatio-temporal proximity of individuals, represented as $R_p$.
% Given $\bar{F}^{idv}$ and bounding box coordinates $P \in \mathbb{R}^{N \times T \times 4}$, we measure the social relation matrix $R$ for social group detection.
% To model accurate social relationships, we investigate the visual similarity and physical relationships among individual. 
We define the visual similarity matrix as $R_s = \mathrm{Softmax} \left ( \textbf{W}_{\theta}\bar{F}^{idv}( \textbf{W}_{\phi} \bar{F}^{idv})^{\top} \right )$, where $\textbf{W}_{\theta}$ and $\textbf{W}_{\phi}$ are trainable matrices.
In addition to the visual similarity, exploiting positional relationships among individuals not just in space but also over time is essential for accurately modeling social relationships.
For this purpose, we introduce the social proximity relation matrix $R_p$, defined with Temporal Generalized IoU (TGIoU), as follows:
\begin{align}
    R_p(i, j) = \textup{TGIoU} (P^i, P^j) = \frac{1}{T} \sum_{t=1}^{T} \textup{GIoU} (p^i_t, p^j_t), \\
    \textsl{where} \: \:
    \textup{GIoU}(p^i_t, p^j_t) = \frac{\left| p^i_t \cap p^j_t \right|}{ \left |p^i_t \cup  p^j_t \right|}  - \frac{ \left| B(p^i_t, p^j_t)\setminus p^i_t \cup p^j_t \right| }{ \left| B(p^i_t, p^j_t) \right|}.
\end{align}
Here, $p_t^i \in [0, 1]^4$ is a vector representing top-left and bottom-right coordinates of a bounding boxes and $B(p^i_t, p^j_t)$ means the smallest box containing both $p^i_t$ and $p^j_t$.
% \begin{align}
%      GIoU(p^i_t, p^j_t) = \left ( \left| p^i_t \cap p^j_t \right| / \left |p^i_t \cup  p^j_t \right|   - \left| B(p^i_t, p^j_t)\setminus p^i_t \cup p^j_t \right| / \left| B(p^i_t, p^j_t) \right| \right)
% \end{align}
Since GIoU~\cite{giou} provides a thorough measure of spatial separation and overlap, TGIoU serves as an informative metric for evaluating physical individual relationships across time.
The final social relation matrix is computed as $R = 1/2 (R_s + R_p)$.

To predict the normalized number of social groups within the scene, we apply a linear layer on the individual features attended by the relation matrix, as $n_g = \textsl{MLP} \left ( \textsl{Avg} (R\bar{F}^{idv}) \right )$.
% To identify social group ids in a given scene, we calculate the normalized number of social groups, represented as $n_g$, by utilizing a linear layer and average pooling on the attended individual features, as follows:
% \begin{align}
%     n_g = \mathrm{MLP} \left ( \mathrm{Avg} (R\bar{F}^{idv}) \right ).
% \end{align}
The actual number of social groups $N_g$ is then calculated by multiplying $N_i$ with $n_g$.
Finally, the social group identifier $\mathcal{G} \in [ 1, N_g]^{N_i}$ are determined via K-means clustering on the relation-augmented features, as follows:
\begin{align}
    \mathcal{G} = \textup{KMeans} \left ( R \bar{F}^{idv}, N_g \right ).
\end{align}
 
% \textbf{Generalized IoU}
% Naive Intersection over Union (IoU) is computed as 

% Generalized IoU~\cite{giou} utilizes the smallest box containing both bounding box and ground truth in the metric to address the error in scenarios without overlapped area. Let $B_o$, $B_A$ and $B_b$ be a grounding truth box, and a predicted box. 
% %https://silhyeonha-git.tistory.com/3
% \textbf{MLP} 

% \textbf{K-means Clustering} 
% K-means clustering algorithm randomly chooses $N_g$ centroids $C = \{ c_1, c_2, ..., c_{N_g} \}$ from the relational values $r^{idv} \in \mathbb{R}^D$. 
% For each value, the algorithm finds the nearest centroid $c_j$ as its corresponding cluster using the sum of squared distance as metric, $D(r_i, c_j) = \sum_{i=1}^{N_g} ||r_i - c_j ||^2$. 
% For each cluster, the algorithm updates its centroid by computing means value along the dimension of data points in the cluster. Finally, it computes the distance between the old and new centroids and repeats the above steps if the overall distance is converged. 

\subsection{Multi-Granular Activity Recognition}
\label{sec:3.3}
For this stage, \tr\ is designed to effectively capture human activity features across multiple granularities by modeling the dependencies among individual, social group, and global contextual information.
As illustrated in Fig.~\ref{fig:mgar}, \tr\ has $L$ layers.
Each layer based on transformer encoder blocks~\cite{transformer} has a dual-path architecture: individual-to-global activity and individual-to-social activity paths.
The first path takes the individual features $\bar{F}^{idv}$ as an input.
To this input, a learnable global token is prepended.
By computing local individual patches and the global token, it facilitates explorations of not just inter-individual interactions among individuals but also global context within the overall scene.
To consider social group activities, the individual features obtained from the first path are grouped based on $\mathcal{G}$ and organized into sequences.
Then, each cluster is prefixed with a learnable social token to capture the dynamics of each social group.
In the individual-to-social activity path, these social tokens are specifically trained to accurately predict their respective social group activities by exploring global-local context from the individual features.
Through multiple layers, \tr\ cooperatively promotes three tasks to create synergistic effects that enhance final activity features of individuals ($\tilde{F}^{idv}\in \mathbb{R}^{N_i \times d}$), social groups ($F^{sg}\in \mathbb{R}^{N_g \times d}$), and global ($F^{glb}\in \mathbb{R}^{d}$).
% $\tilde{F}^{idv}, F^{sg}, f^{glb} = \textup{MGATr}(\bar{F}^{idv}, \mathcal{G})$.

\subsection{Training}
For training \ournet, we use a combination of binary cross entropy and $L2$ loss functions, defined as follows:
\begin{align}
    \mathcal{L} = \mathcal{L}_{idv} + \mathcal{L}_{R} + \mathcal{L}_{aux} + \lambda_{sg} \mathcal{L}_{sg} + \lambda_{glb} \mathcal{L}_{glb} + + \lambda_{n} \mathcal{L}_n ,
\end{align} 
where $\lambda$ represents the balancing hyperparameters.
Here, $\mathcal{L}_{idv}$ indicates individual action loss function, $\mathcal{L}_{R}$ is the loss function for the social relation matrix in group detection, and $\mathcal{L}_{aux}$ is an auxiliary loss for individual action recognition based on $\bar{F}^{idv}$.
$\mathcal{L}_{sg}$ and $ \mathcal{L}_{glb}$ indicate the losses for the social group and global activity recognition, respectively.
% $\mathcal{L}_{R}$ is the loss function for the relation matrix in group detection, and $\mathcal{L}_{aux}$ is an auxiliary loss for individual action recognition based on $\bar{F}^{idv}$.
We apply $L2$ loss for $\mathcal{L}_n$ term, which addresses the estimation of the number of social groups, and use binary cross-entropy loss for the rest.
% While binary cross-entropy loss is utilized for other losses, the loss function for the number of social groups $\mathcal{L}_n$ is based on $L2$ loss.
The ratio of the losses is $\lambda_{sg} :\lambda_{glb} : \lambda_{n} = 3 : 2 : 5$.

\section{Experiments}
\subsection{Dataset and Metrics}
We evaluated \ournet\ on JRDB-PAR dataset~\cite{par_eccv}.
This dataset is an extension of JRDB~\cite{jrdb} and JRDB-act~\cite{jrdb_act} datasets, which contain multi-person scenes in crowded environments captured by a mobile robot.
It consists of 27 videos, of which 20 are for training purposes and 7 are for evaluation.
The dataset includes 27 individual action classes (\eg, \textit{skating} and \textit{holding something}), 11 categories of social group activities (\eg, \textit{sitting closely} and \textit{working together}), and 7 categories of global activities (\eg, \textit{commuting} and \textit{conversing}).
JRDB-PAR contains 27,920 frames with over 628k human bounding boxes.
 
Following~\cite{par_eccv}, we adopted three evaluation metrics for human activity recognition: precision, recall, and $F_1$ scores for activity prediction, denoted as $\mathcal{P}$, $\mathcal{R}$, and $\mathcal{F}$ respectively.
The overall metric for human activity recognition is represented by $\mathcal{F}_a = (\mathcal{F}_i + \mathcal{F}_p + \mathcal{F}_g)/3$.
To evaluate the results of social group detection, we also follow the protocols of~\cite{par_eccv}, using the classical Half metric~\cite{half_metric} (IoU@0.5), the Area Under the Curve (IoU@AUC), and the matrix IoU score (Mat.IoU).
% The learning rate is decayed with the ratio 0.1 after 60 epochs.

\subsection{Implementation Details}
We utilized Inception-v3~\cite{inception} as a frozen backbone pretrained on Collective Activity dataset~\cite{cad}, following~\cite{par_eccv}.
We used 4 \tr\ layers with 4 attention heads and 256 channels.
The size of the input frame is $480 \times 3760$, and the video length $T$ is set to 3.
We implemented the proposed network using 4 NVIDIA GTX 3090 GPUs with PyTorch framework~\cite{pytorch}.
The batch size of each GPU is set to 4.
For training, we used Adam optimizer~\cite{adam} for 60 epochs.
During the initial 15 epochs, we applied a linear warm-up strategy for the learning rate.
For the remainder of the training period, we maintained a constant learning rate of $4 \times 10^{-5}$ with the weight decay of $10^{-2}$.

\subsection{Ablations}
In this section, we conducted extensive experiments to demonstrate the effectiveness of the proposed method.
We evaluated the experiments for Individual Action Recognition (IAR), Social Group Activity Recognition (SGAR), GloBal Activity Recognition (GBAR), and Social Group Detection (SGDet).
Please refer to Sec. \textcolor{red}{A} in the supplementary material for more experiments and analyses.

%======================================================================
\begin{table}[tb]
\caption{Ablation experiments on the Panoramic Positional Embedding (PPE) in the proximity-based relation encoding w.r.t self-attention operation. `S` and `T` indicate spatial and temporal attention, respectively.}
\vspace{-0.4cm}
\label{tab:pano_pos_embed}
\centering
{\small
\begin{tabular}{C{1.4cm} C{1.2cm}|C{.8cm} C{.8cm} C{.8cm} | C{.8cm} C{.8cm} C{.8cm} | C{.8cm} C{.8cm} C{.8cm} | C{1.0cm}}
\hline
\multirow{2}{*}{Attention} & \multirow{2}{*}{PPE} & \multicolumn{3}{c|}{Individual Action} & \multicolumn{3}{c|}{Social Activity} & \multicolumn{3}{c|}{Global Activity} & Overall \\
 & & $\mathcal{P}_i$ & $\mathcal{R}_i$ & $\mathcal{F}_i$ & $\mathcal{P}_p$ & $\mathcal{R}_p$ & $\mathcal{F}_p$ & $\mathcal{P}_g$ & $\mathcal{R}_g$ & $\mathcal{F}_g$ & $\mathcal{F}_a$ \\ \hline
\multirow{2}{*}{T} & \xmark &   51.4&  45.1&  45.7&  32.2&  31.4&  30.5&  58.8&  48.0&  51.4&  42.5\\
 & \cmark &  55.7&  49.8&  50.1&  34.3&  34.6&  33.0&  58.2&  42.8&  47.8&  43.6\\\cdashline{1-12}
\multirow{2}{*}{S} &  \xmark&  55.4&  47.7&  48.9&  31.9&  33.2&  31.2&  58.3&  45.3&  49.5&  43.2\\ 
& \cmark &  56.3&  51.4&  51.2&  34.4&  35.4&  33.4&  60.4&  48.9&  52.6&  45.7\\ \cdashline{1-12}
\multirow{2}{*}{S+T} &  \xmark&  56.2&  50.9&  50.7&  31.6&  33.2&  30.9&  56.7&  44.8&  48.7&  43.4\\ %\cdashline{1-12}
& \cmark & \textbf{59.4}&\textbf{49.7}& \textbf{51.8}& \textbf{36.5}& \textbf{34.7}& \textbf{34.1}& \textbf{63.4}& \textbf{48.8}& \textbf{53.5}& \textbf{46.5}\\ \hline
\end{tabular}
}
\vspace{-0.3cm}
% \vspace{\belowtabcapmargin}
% \setlength{\belowtabcapmargin}{-6pt}
\end{table}
%======================================================================

%======================================================================
\begin{table}[t]
\caption{Ablation experiments on measuring spatial and temporal social proximity. `S` and `T` indicate spatial and temporal axes, respectively.}
\vspace{-0.4cm}
\centering
\begin{tabular}{C{1.5cm} C{1.1cm}|C{1cm} C{1cm} C{1cm} |C{1.5cm} C{1.5cm} C{1.5cm}}
\hline
\multirow{2}{*}{Distance} & \multirow{2}{*}{Axis} & \multicolumn{3}{c|}{Activity Recognition} & \multicolumn{3}{c}{Social Group Detection} \\
& & $\mathcal{F}_i$ & $\mathcal{F}_p$ & $\mathcal{F}_g$ & IoU@0.5 & IoU@AUC&  Mat.IoU \\ \hline
Euclidean &S  & 43.6& 20.7& 52.4& 43.4& 29.6& 26.5\\
Euclidean &S+T& 48.2& 25.7& 52.5& 44.7& 31.4& 26.0\\ \cdashline{1-8  }
GIoU &S& 43.7& 28.4& \textbf{57.0}& 48.7& 37.1& 28.1\\
TGIoU &S+T& \textbf{51.8}& \textbf{34.2}& 53.5& \textbf{56.4} & \textbf{42.5} & \textbf{34.3}\\ \hline
\end{tabular}
% \vspace{-0.3cm}
\label{tab:pano_det}
\vspace{\belowtabcapmargin}
\end{table}
%======================================================================
\vspace{\subsubsecmargin}\subsubsection{Spatio-Temporal Proximity in PAR.}
We investigate the effects of spatio-temporal positional relationships in the PAR task.
Specifically, we validate the impact of spatio-temporal proximity in the proximity-based relation encoding stage through ablation experiments on the panoramic positional embedding and TGIoU.
In Table~\ref{tab:pano_pos_embed}, we summarize the results of \ournet\ with and without panoramic positional embedding in spatial and temporal axes.
The results show that the panoramic positional embedding outperforms conventional embedding in all dimension experiments in terms of all metrics, including precision, recall, and F1 score.
Particularly, compared to using the panoramic positional embedding in either spatial or temporal dimension, applying it in both spatial and temporal dimensions yields improvements of 0.8\% and 2.9\%, respectively, in terms of $F_a$.

Table~\ref{tab:pano_det} shows the experimental results of using spatial and spatio-temporal Euclidean, GIoU, and TGIoU as a proximity metric to demonstrate the impact of the spatio-temporal proximity for PAR.
For spatio-temporal Euclidean distance, we average the spatial Euclidean distances across frames.
From the results, it is evident that spatio-temporal proximity (S+T) outperforms spatial proximity (S).
Specifically, employing Euclidean distance in the spatio-temporal axis leads to a 1.3\% improvement in IoU@0.5 compared to using Euclidean distance. 
Similary, TGIoU yields 56.4\% in terms of IoU@0.5, while GIoU achieves 48.7\%.
Moreover, we see that TGIoU achieves superior performance than the spatio-temporal Euclidean distance in both social group detection and activity recognition.
These results indicate that incorporating TGIoU enables \ournet\ to correctly understand social dynamics in crowded wide scenes.

%======================================================================
\begin{table}[t]
\caption{Ablation study on the similarity relation $R_s$ and the physical relation $R_p$.}
\vspace{-0.4cm}
\label{tab:R}
\centering
\begin{tabular}{C{0.8cm} C{0.8cm}|C{1cm} C{1cm} C{1cm} |C{1.5cm} C{1.5cm} C{1.5cm}}
\hline
\multirow{2}{*}{$R_s$} & \multirow{2}{*}{$R_p$ } & \multicolumn{3}{c|}{Activity Recognition} & \multicolumn{3}{c}{Social Group Detection} \\
 & & $\mathcal{F}_i$ & $\mathcal{F}_p$ & $\mathcal{F}_g$ & IoU@0.5 & IoU@AUC&  Mat.IoU \\ \hline
& & 46.2& 22.7& 53.1& 32.2& 20.6& 17.9\\
\checkmark & & 48.1& 26.1& 52.2& 37.6& 24.6& 21.9\\
& \checkmark &51.3& 33.2& 48.8& 55.8& \textbf{44.0}& 33.9\\ \cdashline{1-8}
\checkmark & \checkmark &\textbf{51.8}& \textbf{34.2}& \textbf{53.5}& \textbf{56.4} & \underline{42.5} & \textbf{34.3}\\ \hline
\end{tabular}
\vspace{-0.3cm}
% \vspace{\belowtabcapmargin}
\end{table}
%----------------------------------------------------------------------

\vspace{\subsubsecmargin}\subsubsection{Social Group Relation.}
To demonstrate the effectiveness of the similarity relation $R_s$ and the social proximity relation $R_p$ in the social relation $R$, we ablate them and the results are presented in Table~\ref{tab:R}.
Compared to the baseline not using any social relation, either utilizing $R_s$ or $R_p$ results in higher group detection performance by 5.4\% and 23.6\% in terms of IoU@0.5, respectively.
Moreover, the performances of human activity recognition are also improved.
Especially, $\mathcal{F}_p$ is increased by 3.4\%p and 10.5\%p with $R_s$ and $R_p$, respectively.
Notably, while $R_s$ is effective, spatio-temporal proximity from $R_p$ is a crucial factor for defining individual relationships within a panoramic scene.
Ultimately, employing both $R_s$ and $R_p$ achieves 56.4\% of IoU@0.5 for SGDet and 34.2\% of $\mathcal{F}_p$ for SGAR.

%======================================================================
\begin{figure}[t]
    \centering
    \includegraphics[width=0.9\textwidth]{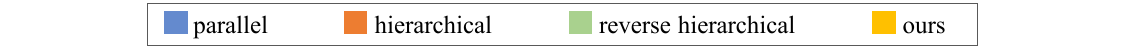}
    {
    \begin{subfigure}[b]{0.3\textwidth}
         \centering
         \includegraphics[width=\textwidth]{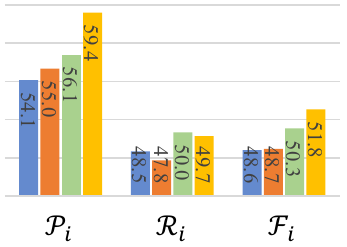}
         \caption{\footnotesize Individual action}
         \label{fig:y equals x}
     \end{subfigure}
     \hfill
     \begin{subfigure}[b]{0.3\textwidth}
         \centering
         \includegraphics[width=\textwidth]{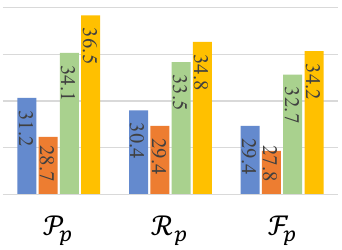}
         \caption{\footnotesize Social group activity}
         \label{fig:three sin x}
     \end{subfigure}
     \hfill
     \begin{subfigure}[b]{0.3\textwidth}
         \centering
         \includegraphics[width=\textwidth]{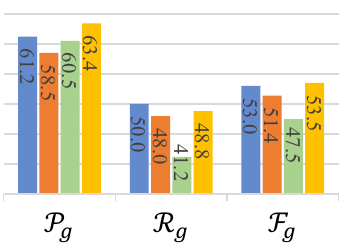}
         \caption{\footnotesize Global activity}
         \label{fig:five over x}
     \end{subfigure}
\vspace{-0.3cm}
\caption{The results of ablation experiments of a dual-path architecture in \tr.}
\label{fig:three graphs}
\vspace{\belowfigcapmargin}
}
\end{figure}
%======================================================================

\vspace{\subsubsecmargin}\subsubsection{Dual-Path Activity Transformer.}
\label{sec:dptr}
%======================================================================
\begin{wrapfigure}{r}{0.45\textwidth}
\vspace{-0.8cm}
 \begin{center}
 \includegraphics[width=0.45\textwidth]{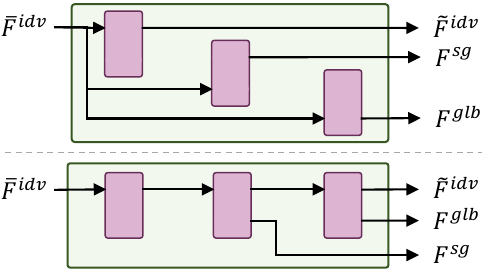}
 \end{center}
 \vspace{-0.6cm}
 \caption{Simple illustration of a parallel (upper) and hierarchical (lower) structures.}
 \label{fig:warpfig}
 \vspace{-0.7cm}
\end{wrapfigure}
%======================================================================
We analyze the effectiveness of a dual-path architecture for multi-granular activity recognition.
For comparisons, we model three types of transformer structure: parallel, hierarchical, and reverse hierarchical.
Each of these models comprises three Transformer encoder blocks to refine features of individual actions, social group activities, and global activities. 
As depicted in Fig.~\ref{fig:warpfig}, the parallel structure independently extracts specific granular activities, while the hierarchical structure sequentially captures activity information from smaller to larger spatial granularity.
The reverse hierarchical structure operates inversely to the hierarchical, from larger to smaller spatial granularity.
% In the parallel structure, the three blocks operate independently to extract features corresponding to specific granular activities.
% The hierarchical structure operates sequentially,  
For details, please refer to Sec. \textcolor{red}{A.1} in the supplementary material.

The performances of \tr\ with different structures are illustrated in Fig.~\ref{fig:three graphs}.
The parallel architecture outperforms the hierarchical structure in both SGAR and GBAR, suggesting that individual action information is essential for achieving high performance in these tasks.
Additionally, the reverse hierarchical structure, which leverages global context for encoding social group activity features, exhibits superior performance in SGAR compared to the parallel and forward hierarchical structures that only use individual information. 
These results indicate the significance of global context in defining social group activity and the mutual influence of individual, social group, and global contexts.
\tr\ outperforms others in all multi-granular activity recognition tasks by creating synergistic effects of the individual-to-global and individual-to-social paths.
From these results, we demonstrate that mutually interacting with contextual understanding of multi-spatial activities is imperative for PAR.
% incorporating both global and local contexts enhances not only social group activity performance but also individual and global activity performance.
%global에서는 어쨌든 individual 정보가 들어가는 ours랑 parallel에서 성능이 잘 나옴
%근데 global context까 빠진 parallel의 social group성능이 낮음.
%오히려 global context가 있는 reverse가 local context만 있는 hierarchical보다 social group성능이 좋음.
%따라서 global-local context를 추출하는 것이 social group activity성능 뿐만아니라 individual과 global 성능까지 함께 높아질 수 있는 뽀인뚜(찡긋)

\subsection{Comparison with the State-of-the-arts}
% \vspace{\subsubsecmargin}
%======================================================================
\begin{table}[t]
\centering
\caption{Comparative results of the panoramic activity recognition (\%). The best scores are marked in \textbf{bold} and the second best ones are \underline{underlined}.}
\vspace{-0.4cm}
{\small
\begin{tabular}{C{2.3cm} | C{.85cm} C{.85cm} C{.85cm} | C{.85cm} C{.85cm} C{.85cm} | C{.85cm} C{.85cm} C{.85cm} | C{1.1cm}}
\hline
\multirow{2}{*}{Method} & \multicolumn{3}{c|}{Individual Action} & \multicolumn{3}{c|}{Social Activity} & \multicolumn{3}{c|}{Global Activity} & Overall \\
 & $\mathcal{P}_i$ & $\mathcal{R}_i$ & $\mathcal{F}_i$ & $\mathcal{P}_p$ & $\mathcal{R}_p$ & $\mathcal{F}_p$ & $\mathcal{P}_g$ & $\mathcal{R}_g$ & $\mathcal{F}_g$ & $\mathcal{F}_a$ \\ \hline
ARG~\cite{arg} & 39.9 & 30.7 & 33.2 & 8.7 & 8.0 & 8.2 & \textbf{63.6} & 44.3 & 50.7 & 30.7 \\
SA-GAT~\cite{sa-gat} & 44.8 & 40.4 & 40.3 & 8.8 & 8.9 & 8.8 & 36.7 & 29.9 & 31.4 & 26.8 \\
JRDB-Base~\cite{jrdb_act} & 19.1 & 34.4 & 23.6 & 14.3 & 12.2 & 12.8 & 44.6 & 46.8 & 45.1 & 27.2 \\
JRDB-PAR~\cite{par_eccv} & 51.0 & 40.5 & 43.4 & 24.7 & 26.0 & 24.8 & 52.8 & 31.8 & 38.8 & 35.6 \\
MUP~\cite{mup} & \underline{55.4} & \underline{44.8}& \underline{47.7} & \underline{25.4} & \underline{26.6} & \underline{25.1} & 58.0 & \textbf{49.0} & \underline{51.8} & \underline{41.5} \\ \cdashline{1-11}
Ours & \textbf{59.4}&\textbf{49.7}& \textbf{51.8}& \textbf{36.5}& \textbf{34.8}& \textbf{34.2}& \underline{63.4}& \underline{48.8}& \textbf{53.5}& \textbf{46.5}\\ \hline
\end{tabular}
}
\vspace{-0.3cm}
\label{tab:main}
\end{table}
%----------------------------------------------------------------------

\subsubsection{Human Activity Recognition.}
In Table~\ref{tab:main}, we compare the proposed method with other comparative methods for PAR.
\ournet\ surpasses the comparative methods in the overall performance.
Particularly, \ournet\ achieves huge performance improvement in social group activity recognition, attaining 36.5\% in $\mathcal{P}_p$, 34.8\% in $\mathcal{R}_p$, and 34.2\% in $\mathcal{F}_p$.
This represents significant improvements of 10\%, 8.2\%, and 9.1\%, respectively, compared to MUP~\cite{mup}, which hierarchically recognizes multi-granular activities.

%======================================================================
\begin{table}[t]
\centering
\caption{Performance comparison with ground truth group detection results for the panoramic activity recognition.}
\vspace{-0.4cm}
{\small
\begin{tabular}{C{2.3cm} | C{.85cm} C{.85cm} C{.85cm} | C{.85cm} C{.85cm} C{.85cm} | C{.85cm} C{.85cm} C{.85cm} | C{1.1cm}}
\hline
\multirow{2}{*}{Method} & \multicolumn{3}{c|}{Individual Action} & \multicolumn{3}{c|}{Social Activity} & \multicolumn{3}{c|}{Global Activity} & Overall \\
 & $\mathcal{P}_i$ & $\mathcal{R}_i$ & $\mathcal{F}_i$ & $\mathcal{P}_p$ & $\mathcal{R}_p$ & $\mathcal{F}_p$ & $\mathcal{P}_g$ & $\mathcal{R}_g$ & $\mathcal{F}_g$ & $\mathcal{F}_a$ \\ \hline
AT~\cite{at} & 38.9 & 33.9 & 34.6 & 32.5 & 32.3 & 32.0 & 21.2 & 19.1 & 19.8 & 28.8 \\
SACRF~\cite{sacrf} & 31.3 & 23.6 & 25.9 & 25.7 & 24.5 & 24.8 & 42.9 & 35.5 & 37.6 & 29.5 \\
Dynamic~\cite{dynamic} & 40.7 & 33.4 & 35.1 & 33.5 & 30.1 & 30.9 & 37.5 & 27.1 & 30.6 & 32.2 \\
HiGCIN~\cite{higcin} & 34.6 & 26.4 & 28.6 & 34.2 & 31.8 & 32.2 & 39.3 & 30.1 & 33.1 & 31.3 \\
ARG~\cite{arg} & 42.7 & 34.7 & 36.6 & 27.4 & 26.1 & 26.2 & 26.9 & 21.5 & 23.3 & 28.8 \\
SA-GAT~\cite{sa-gat} & 39.6 & 34.5 & 35.0 & 32.5 & 32.5 & 30.7 & 28.6 & 24.0 & 25.5 & 30.4 \\
JRDB-Base~\cite{jrdb_act} & 21.5 & 44.9 & 27.7 & 54.3 & 45.9 & 48.5 & 38.4 & 33.1 & 34.8 & 37.0 \\
JRDB-PAR~\cite{par_eccv} & 54.3 & 44.2 & 46.9 & 50.3 & 52.5 & 50.1 & 42.1 & 24.5 & 30.3 & 42.4 \\ 
MUP~\cite{mup} & 56.8& 45.6 & 48.6 & 55.7 & 49.7 & 51.3 & 57.0 & 46.2 & 47.3 & 49.2 \\ \cdashline{1-11}
Ours & \textbf{60.4}& \textbf{50.5}& \textbf{52.7}& \textbf{56.5}& \textbf{54.8}& \textbf{53.5}& \textbf{62.9}& \textbf{48.4}& \textbf{53.1}& \textbf{53.1}\\ \hline
\end{tabular}
}
\vspace{-0.3cm}
\label{tab:main_gt}
\end{table}
%----------------------------------------------------------------------

In Table~\ref{tab:main_gt}, we further compare the proposed \ournet\ with other state-of-the-art methods using ground-truth social group detection.
This comparison is intended to solely evaluate the capability of recognizing multi-granular activity.
Both the compared methods and \ournet\ achieve enhanced performance in social activity detection when utilizing ground-truth social group detection. 
Specifically, \ournet\ with the ground-truth group detection results achieves 19.3\% improvement in $\mathcal{F}_p$ compared to not using it.
Notably, \ournet consistently outperforms the comparison methods even when utilizing ground-truth social group detection results.
This shows the superiority of the proposed method in capturing the dependencies among multi-spatial granular activities and accurately predicting them.

\vspace{\subsubsecmargin}\subsubsection{Social Group Detection.}
\label{sec:sgd}
We evaluate \ournet\ against state-of-the-art methods for SGDet as presented in Table~\ref{tab:main_det}.
Compared to the other methods~\cite{arg,sa-gat,jrdb_act,par_eccv}, \ournet\ achieves the best performances of 56.4\% in IoU@0.5, 42.5\% in IoU@AUC, and 34.3\% in Mat.IoU.
By the enhancements in social group detection performance, \ournet\ achieves improvements not only in $F_p$ but also in $F_i$ and $F_g$.
We further evaluate \ournet\ with the ground-truth number of social groups $N_g$, denoted as `Ours*' in Table~\ref{tab:main_det}.
We note that the utilization of ground-truth $N_g$ in \ournet\ enhances the performances of human activity recognition and social group detection, particularly in SGAR.

%======================================================================
\begin{table}[t]
\centering
\caption{Performance comparisons of group detection (\%). * indicates that we use the ground-truth number of social groups as input. The best scores are marked in \textbf{bold} and the second best ones are \underline{underlined}.}
\vspace{-0.4cm}
{\small
\begin{tabular}{C{3cm}|C{1cm} C{1cm} C{1cm} |C{1.7cm} C{1.7cm} C{1.7cm}}
\hline
\multirow{2}{*}{Method} & \multicolumn{3}{c|}{Activity Recognition} & \multicolumn{3}{c}{Social Group Detection} \\
 & $\mathcal{F}_i$ & $\mathcal{F}_p$ & $\mathcal{F}_g$ & IoU@0.5 & IoU@AUC & Mat. IoU \\ \hline
ARG~\cite{arg} & 33.2 & 8.2 & 50.7 & 35.2& 21.6& 19.3\\
SA-GAT~\cite{sa-gat} & 40.3 & 8.8 & 31.4 & 29.1 & 20.4 & 16.6 \\
JRDB-Base~\cite{jrdb_act} & 23.6 & 12.8 & 45.1 & 38.4 & 26.3 & 20.6 \\
JRDB-PAR~\cite{par_eccv} & 43.4 & 24.8 & 38.8 & 53.9 & 38.1 & 30.6 \\ \cdashline{1-7}
Ours  & \underline{51.8}& \underline{34.2}& \textbf{53.5}& \underline{56.4}& \underline{42.5} & \underline{34.3}\\
Ours* & \textbf{52.2}& \textbf{35.8}& \underline{52.9}& \textbf{59.7}& \textbf{45.5}& \textbf{40.2}\\ \hline
\end{tabular}
}
\vspace{-0.2cm}
\label{tab:main_det}
\end{table}
%----------------------------------------------------------------------

%======================================================================
\begin{figure}[t]
    {
    \begin{subfigure}[b]{0.49\textwidth}
         \centering
         \includegraphics[width=\textwidth]{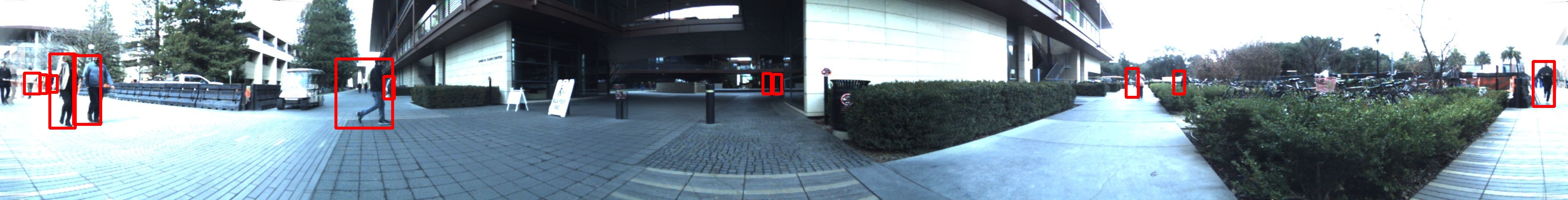}
         \vspace{-0.3cm}
         \label{fig:relation_matrix_a}
     \end{subfigure}
     \begin{subfigure}[b]{0.49\textwidth}
         \centering
         \includegraphics[width=\textwidth]{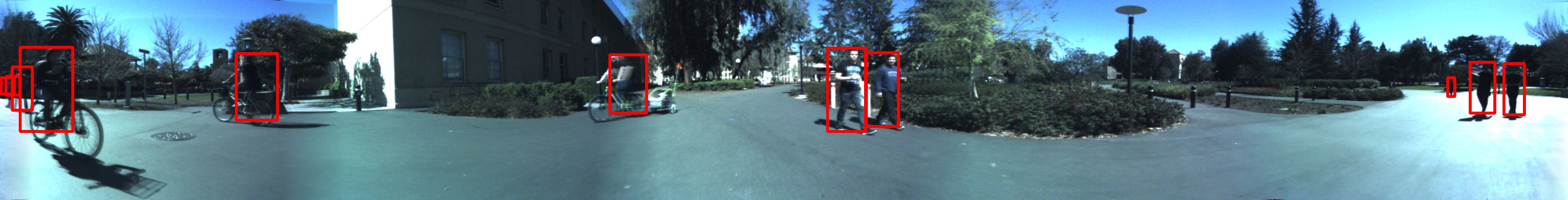}
         \vspace{-0.3cm}
         \label{fig:relation_matrix_b}
     \end{subfigure}
     % \vspace{-0.3cm}
     }
     {
    \begin{subfigure}[b]{0.49\textwidth}
         \centering
         \includegraphics[width=\textwidth]{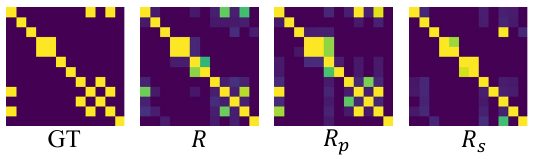}
         \caption{}
         \label{fig:relation_matrix_a}
     \end{subfigure}
     \hfill
     \begin{subfigure}[b]{0.49\textwidth}
         \centering
         \includegraphics[width=\textwidth]{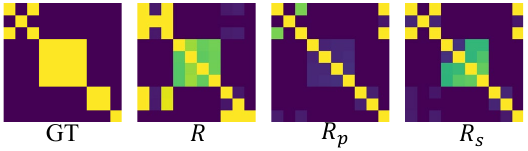}
         \caption{}
         \label{fig:relation_matrix_b}
     \end{subfigure}
     \vspace{-0.3cm}
     }
    \caption{Visualization of the ground-truth (GT) and predicted relation matrix $R$, the proximity relation matrix $R_p$, and the similarity matrix $R_s$. Best viewed zoomed in on screen.}
    \label{fig:relation_matrix}
    \vspace{\belowfigcapmargin}
\end{figure}
%----------------------------------------------------------------------
\subsection{Visualization}
% \vspace{\subsubsecmargin}
\subsubsection{Relation Matrix.}
In Fig.~\ref{fig:relation_matrix}, we visualize the relation matrix $R$, the social proximity relation matrix $R_p$, and the similarity matrix $R_s$, which are $N_i \times N_i$ matrices.
The ground-truth social relation matrix takes a value of 1 for individuals belonging to the same social group and 0 for otherwise.
% The yellowish color indicates a high degree of social relation, while darker colors signify the absence of social relation.
In Fig.~\ref{fig:relation_matrix_a}, it is evident that $R_p$ closely aligns with the ground-truth social relation compared to $R_s$.
This emphasizes the significant contribution of the spatio-temporal proximity ($R_s$) for discerning social dynamics in a crowded scene.
Conversely, in scenarios depicted in Fig.~\ref{fig:relation_matrix_b}, where bounding boxes are relatively large and fewer people are present, we observe that the visual similarity ($R_s$) is effective.
For additional results, please see Sec. \textcolor{red}{B.1} in the supplementary material.
% Compared to Fig.~\ref{fig:relation_matrix_a}, which has small bounding boxes in a crowded scene, visual similarity ($R_S$) is effective in scenarios where bounding boxes are relatively large with fewer people (see Fig.~\ref{fig:relation_matrix_b}).

%======================================================================
\begin{figure}
    \centering 
    \begin{subfigure}[t]{0.95\textwidth}
        \centering
        \includegraphics[width=\textwidth]{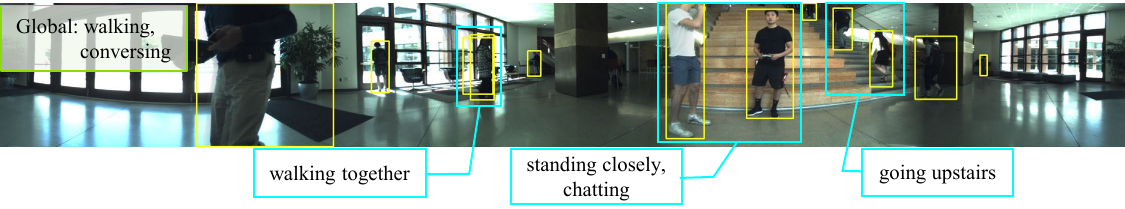}
        \caption{GT}
    \end{subfigure}
    
    \begin{subfigure}[t]{0.95\textwidth}
        \includegraphics[width=\textwidth]{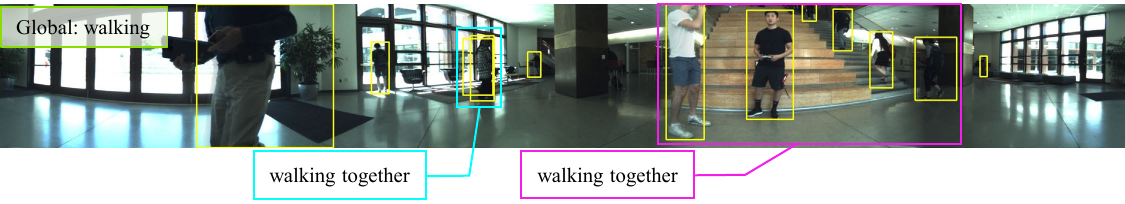}
        \caption{without the proximity relation matrix $R_p$}
    \end{subfigure}
    \begin{subfigure}[t]{0.95\textwidth}
        \includegraphics[width=\textwidth]{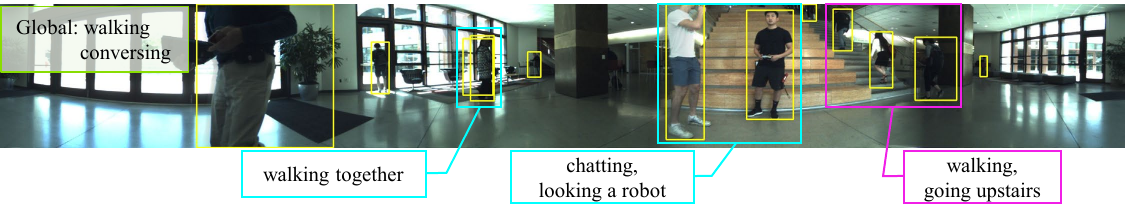}
        \caption{\ournet(Ours)}
    \end{subfigure}
    \vspace{-0.3cm}
    \caption{The visual comparison of the social group activity detection and global activity recognition on JRDB-PAR dataset~\cite{par_eccv}.
    % (a) Ground-truth, (b) \ournet\ without the proximity relation $R_p$, and (c) \ournet.
    The individual bounding boxes are marked in \textcolor{yellow!40!orange}{yellow}. Misclassified social group bounding boxes are indicated in \textcolor{magenta}{magenta}, while ground-truth and correctly predicted bounding boxes are in \textcolor{cyan}{cyan}.}
    \label{fig:result}
    \vspace{\belowfigcapmargin}
\end{figure}
%----------------------------------------------------------------------
\vspace{\subsubsecmargin}\subsubsection{Qualitative Result.}
Figure~\ref{fig:result} shows the visual comparisons between the ground-truth, \ournet\ with and without the social proximity relation $R_p$ on JRDB-PAR dataset~\cite{par_eccv}.
We observe that not utilizing $R_p$ results in inaccurate or missed social group detections.
The spatio-temporal positional information enables \ournet\ to infer that the two men standing in front of the stairs continue to converse while the three individuals on the stairs are still ascending, indicating no social relationship between those two groups. 
However, the person at the right bottom of the stairs, despite not engaging in any social group interaction with others, is misclassified as part of a social group due to exhibiting similar behavior and physical closeness.
For more visualizations, please refer to Sec. \textcolor{red}{B.2} in the supplementary material.

\section{Conclusion}
In this paper, we have proposed a novel network for Panoramic Activity Recognition (PAR), named Social Proximity-aware Dual-Path Network (\ournet).
By incorporating spatio-temporal positional relationships among individuals throughout the panoramic positional embedding and spatio-temporal proximity relations, \ournet\ accurately captures social and global dynamics within a crowded panoramic scene.
Furthermore, we have introduced Dual-Path Activity 
Transformer (\tr), which consists of individual-to-global and individual-to-social paths.
DPATr synergistically enhances final predictions by mutually reinforcing contextual understandings of multi-spatial activities.
In this end, the proposed \ournet\ sets new state-of-the-art records for PAR.
We discuss limitations in Sec. \textcolor{red}{C} of the supplementary material.
% Throughout extensive experiments, we showed  that \ournet\ is an effective method for PAR, achieving new state-of-the-art performance.

% acknowledgement
% \hfill
% \noindent\textbf{Acknowledgment}This work was conducted by Center for Applied Research in Artificial Intelligence (CARAI) grant funded by DAPA and ADD (UD230017TD). 


\clearpage
\bibliographystyle{splncs04}
\bibliography{egbib}

\clearpage

\renewcommand*{\thesection}{\Alph{section}}

\title{Appendix} 


\author{ }


\institute{}

\maketitle

%======================================================================
\begin{figure}[h]
\centering
    {
    \begin{subfigure}[b]{0.48\textwidth}
         \centering
         \includegraphics[width=\textwidth]{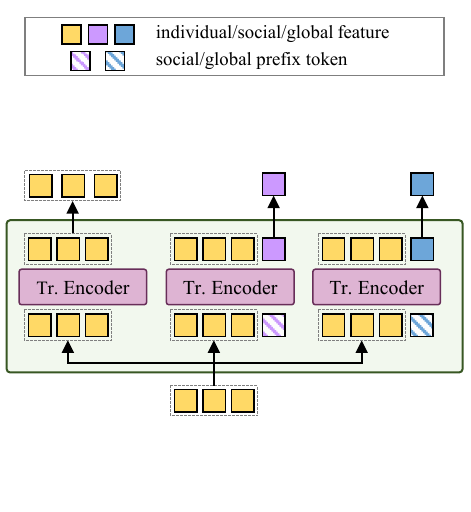}
         \caption{}
         \label{fig:dual_parallel}
     \end{subfigure}
     \begin{subfigure}[b]{0.22\textwidth}
         \centering
         \includegraphics[width=\textwidth]{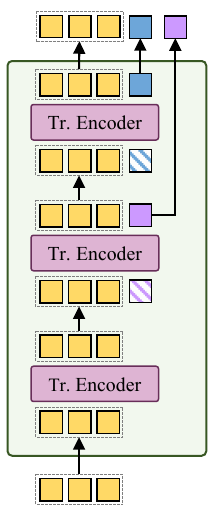}
         \caption{}
         % \vspace{-0.3cm}
         \label{fig:dual_hier}
     \end{subfigure}
    \begin{subfigure}[b]{0.22\textwidth}
         \centering
         \includegraphics[width=\textwidth]{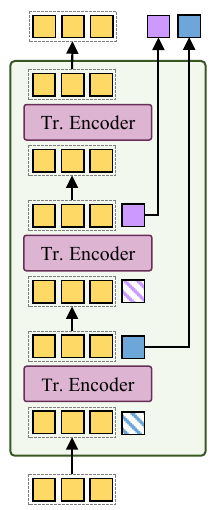}
         \caption{}
         \label{fig:dual_reverse}
         % \vspace{-0.3cm}
     \end{subfigure}
     }
    \caption{A detailed overview of (a) parallel, (b) hierarchical, and (c) reverse hierarchical architectures. }
    \label{fig:dual}
    \vspace{\belowfigcapmargin}
\end{figure}
%----------------------------------------------------------------------

\section{Ablation Experiments}
\label{supp:ablation}
In this section, we conducted additional experiments to demonstrate the effectiveness of the proposed method.
We evaluated the experiments for Individual Action Recognition (IAR), Social Group Activity Recognition (SGAR), GloBal Activity Recognition (GBAR), and Social Group Detection (SGDet).

\subsection{Dual-Path Activity Transformer}
In Sec. \textcolor{red}{4.3} in the manuscript, we design three types of transformer structures to compare with the proposed Dual-Path Activity Transformer (DPATr).
Figure~\ref{fig:dual} illustrates the detailed mechanism of the ablated architectures: parallel, hierarchical, and reverse hierarchical.
Each of these models consists of three transformer encoder blocks~\cite{transformer} dedicated to enhancing features of individual actions, social group activities, and global activities, respectively. 
In the parallel architecture, these blocks operate independently to capture features related to specific granular activities from the self-attended individual features $\bar{F}^{idv}$. 
The hierarchical structure sequentially extracts activity information from smaller to larger spatial granularity. 
In contrast, the reverse hierarchical structure operates conversely, extracting activity information from larger to smaller spatial granularity.
As illustrated in Fig.~\textcolor{red}{3.a} in the manuscript, each \tr\ layer comprises an individual-to-global path and an individual-to-social path.
In the individual-to-social path, richer social-level representations are extracted by leveraging the global-local context explored in the individual-to-global path. 
By mutually reinforcing contextual understanding of multi-spatial activities through multiple layers, \ournet\ achieves the most superior performances across all metrics (see Sec. \textcolor{red}{4.3} in the manuscript).

\subsection{Spatio-Temporal Individual Attention.}
We ablate three attentions across temporal, height, and width axes in the proximity-based relation encoding.
The results are shown in Table~\ref{tab:idv_attn}.
We observe that \ournet\ with either the temporal attention $A^t$ or the spatial attention (\ie, $A^h$ and $A^w$) improves the performances by exploiting informative action features of each individual.
Incorporating either $A^h$ or $A^w$ with $A^t$ improves the performance of PAR, particularly IAR and GBAR. 
Specifically, compared to solely using $A^t$, a combination of $A^t$ and $A^h$ improves the performances of IAR and GBAR by 3.5\% and 2.6\% in F1 score, respectively.
Similarly, using $A^t$ and $A^w$ achieves 3.3\% improvement of SGAR and 1.6\% improvement of GBAR, in terms of F1 score.
By applying both spatial and temporal attentions, our \ournet\ achieves the best overall performance, resulting 46.5\%.

%======================================================================
\begin{table}[t]
\centering
\caption{Ablation experiments on the spatio-temporal individual self-attention. $A^t$, $A^h$, and $A^w$ indicate attentions along temporal, height, and width dimensions, respectively. The best scores are marked in \textbf{bold} and the second best ones are \underline{underlined}.}
\vspace{-0.2cm}
\small
\begin{tabular}{C{.6cm} C{.6cm} C{.6cm} | C{.8cm} C{.8cm} C{.8cm} | C{.8cm} C{.8cm} C{.8cm} | C{.8cm} C{.8cm} C{.8cm} | C{1.2cm}}
\hline
\multirow{2}{*}{$A^t$} & \multirow{2}{*}{$A^h$} & \multirow{2}{*}{$A^w$} &\multicolumn{3}{c|}{Individual Action} & \multicolumn{3}{c|}{Social Activity} & \multicolumn{3}{c|}{Global Activity} & Overall \\ 
 & & & $\mathcal{P}_i$ & $\mathcal{R}_i$ & $\mathcal{F}_i$ & $\mathcal{P}_p$ & $\mathcal{R}_p$ & $\mathcal{F}_p$ & $\mathcal{P}_g$ & $\mathcal{R}_g$ & $\mathcal{F}_g$ & $\mathcal{F}_a$ \\ \hline
 &  &  & 51.9 & 38.9 & 42.7 & 29.8 & 23.9 & 25.6 & 54.0& 39.2 & 44.0 & 37.4\\
 & \checkmark &\checkmark & 52.4 & 46.7 & 47.0 & 30.6 & 30.4 & 29.2 & 56.2 & 44.8 & 48.3 & 41.5 \\
 \checkmark &  &  & 56.4 & 44.1 & 47.5 & 29.6 & 28.5 & 27.8 & 54.2 & 42.1 & 45.6 & 40.3 \\
\checkmark & \checkmark &  & 57.8 & 49.5 & 51.0 & 31.3 & 30.2 & 29.6 & 59.2& 42.8 &48.2 & 42.9\\
\checkmark &  & \checkmark& 57.5 & \textbf{50.4} & 51.4 & 32.3 & 32.4 & 31.1 & 58.0 & 42.5 & 47.2 & 43.2\\ \cdashline{1-13}
\checkmark & \checkmark  & \checkmark & \textbf{59.4} & \underline{49.7} & \textbf{51.8}  & \textbf{36.5} & \textbf{34.8} & \textbf{34.2} & \textbf{63.4} & \textbf{48.8} & \textbf{53.5} & \textbf{46.5} \\ \hline
\end{tabular}
\label{tab:idv_attn}
\end{table}
%----------------------------------------------------------------------

%======================================================================
\begin{table}[t]
\centering
\caption{Ablation experiments on clustering algorithm, spectral clustering and K-means clustering. The best scores are marked in \textbf{bold}.}
\vspace{-0.2cm}
% \begin{tabular}{C{2cm}|C{0.8cm} C{0.8cm} C{0.8cm} |C{1.6cm} C{1.6cm} C{1.6cm}}
% \hline
% clustering                                     & $\mathcal{F}_i$ & $\mathcal{F}_p$ & $\mathcal{F}_g$ & IoU@0.5 & IoU@AUC & Mat.IoU \\ \hline
% Spectral~\cite{spectral} &       52.5&       30.5&       53.2&         49.1&         34.8&         27.7\\
% K-means                                        &       51.8&       34.2&       53.2&         56.4&         42.5&         34.4\\ \hline
% \end{tabular}
\begin{tabular}{C{2cm}|C{1.6cm} C{1.6cm} C{1.6cm}}
\hline
clustering               & IoU@0.5 & IoU@AUC & Mat.IoU \\ \hline
Spectral~\cite{spectral} &     49.1&         34.8&         27.7\\
K-means                  &     \textbf{56.4}&         \textbf{42.5}&         \textbf{34.4}\\ \hline
\end{tabular}

\label{tab:clustering}
\end{table}
%----------------------------------------------------------------------

\subsection{Clustering Algorithms}
Compared with previous works~\cite{par_eccv,mup} employing a graph-based Spectral clustering~\cite{spectral}, we utilize a parametric-based clustering scheme with the predicted number of the social groups.
Table~\ref{tab:clustering} shows the results of \ournet\ with Spectral clustering and K-means clustering, which is a parameter-based method.
\ournet\ using K-means clustering outperforms using Spectral clustering in SGDet.
In particular, K-means clustering demonstrates performance improvements across various metrics: achieving 56.4\% in IoU@0.5, 42.5\% in IoU@AUC, and 34.4\% in Mat.IoU. 
These results signify enhancements of 7.3\%, 7.7\%, and 6.7\%, respectively.
Spectral clustering encounters challenges in determining the optimal cluster number due to its sensitivity to the kernel function. 
Additionally, it may face scalability and stability issues.
For these reasons, K-means clustering, which utilizes the predicted number of clusters, exhibits greater robustness than spectral clustering in social group activity detection in a crowded scene.

%======================================================================
\begin{table}[t]
\centering
\caption{Ablation experiments on the auxiliary loss $\mathcal{L}_{aux}$ and the relation loss $\mathcal{L}_R$ functions. The best scores are marked in \textbf{bold}.}
\vspace{-0.2cm}
{\small
\begin{tabular}{C{1cm} C{1cm} | C{.8cm} C{.8cm} C{.8cm} | C{.8cm} C{.8cm} C{.8cm} | C{.8cm} C{.8cm} C{.8cm} | C{1.2cm}}
\hline
 \multirow{2}{*}{$L_{aux}$} & \multirow{2}{*}{$L_{R}$} & \multicolumn{3}{c|}{Individual Action} & \multicolumn{3}{c|}{Social Activity} & \multicolumn{3}{c|}{Global Activity} &  \multirow{2}{*}{IoU@0.5} \\
 &  & $\mathcal{P}_i$ & $\mathcal{R}_i$ & $\mathcal{F}_i$ & $\mathcal{P}_p$ & $\mathcal{R}_p$ & $\mathcal{F}_p$ & $\mathcal{P}_g$ & $\mathcal{R}_g$ & $\mathcal{F}_g$ &  \\ \hline
 & & 50.7 & 45.5& 45.5& 29.4& 29.7 &28.3& 58.6 &43.3 & 48.4 & 51.8 \\
  & \checkmark & 53.2 & 48.7 &47.8 & 30.4 & 29.9 & 28.9 & 59.9 & 43.9 & 49.2 & 51.4 \\ 
  \checkmark &  & 56.7 & 47.3 & 49.4 & 33.9 & 31.0 & 31.2 & 60.4 & 47.7 & 51.6 & 53.3 \\ \cdashline{1-12}
 \checkmark & \checkmark & \textbf{59.4} & \textbf{49.7} & \textbf{51.8}  & \textbf{36.5} & \textbf{34.8} & \textbf{34.2} & \textbf{63.4} & \textbf{48.8} & \textbf{53.5} & \textbf{56.4}\\ \hline
\end{tabular}
}

\label{tab:loss}
\end{table}
%----------------------------------------------------------------------
%======================================================================
\begin{table}[t]
\caption{Ablation experiments on the balancing parameters between loss functions.$\lambda_{idv}$, $\lambda_{R}$, and $\lambda_{aux}$ represents weights of the individual action loss, the relation loss, and the auxiliary loss functions, respectively}
\vspace{-0.2cm}
\centering
\begin{tabular}{C{2.3cm}|C{1cm} C{1cm}C{1cm}C{1cm}|C{1.3cm}C{1.3cm}C{1.3cm}}
\hline
\multirow{2}{*}{$\lambda_{idv}:\lambda_{R}:\lambda_{aux}$}  & \multicolumn{4}{c|}{Activity Recognition} & \multicolumn{3}{c}{Social Group Detection} \\
& $\mathcal{F}_i$ & $\mathcal{F}_p$ & $\mathcal{F}_g$ & $\mathcal{F}_a$ & IoU@0.5 & IoU@AUC & Mat.IoU\\ \hline
1 : 1 : 3&  53.1&  30.4&  55.2&  46.2&  50.0&  35.8&  28.1\\
1 : 3 : 1&  50.0&  32.1&  52.6&  44.9&  56.1&  42.0&  34.6\\ 
3 : 1 : 1&  52.5&  29.8&  52.9&  45.1&  51.7&  39.3&  29.8\\ \cdashline{1-8}
1 : 1 : 1&  51.8&  34.2&  53.5&  46.5&  56.4&  42.5&  34.3\\ \hline
\end{tabular}
\label{tab:lambda}
\end{table}
%----------------------------------------------------------------------
%======================================================================
\begin{figure}[t]
    {
    \begin{subfigure}[b]{0.49\textwidth}
         \centering
         \includegraphics[width=\textwidth]{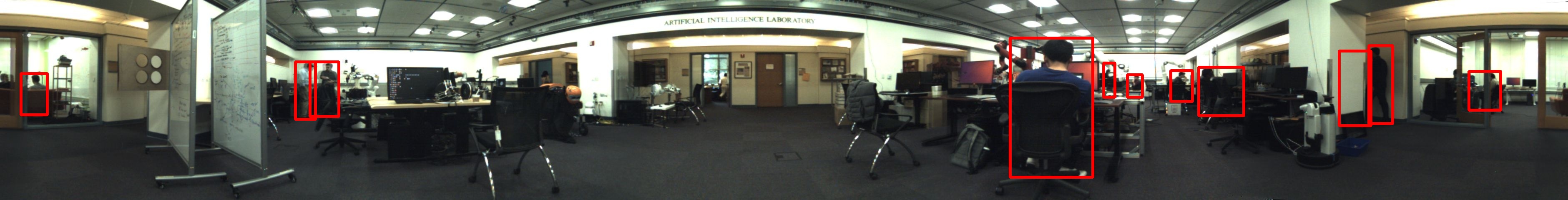}
         \vspace{-0.3cm}
     \end{subfigure}
     \begin{subfigure}[b]{0.49\textwidth}
         \centering
         \includegraphics[width=\textwidth]{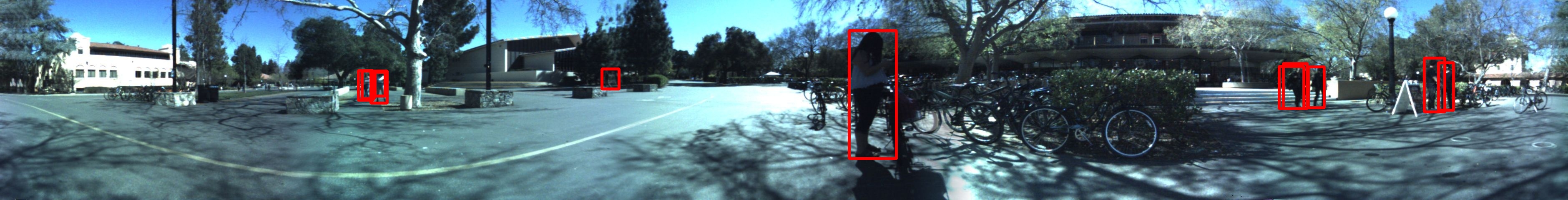}
         \vspace{-0.3cm}
     \end{subfigure}
     % \vspace{-0.3cm}
     }
    {
    \begin{subfigure}[b]{0.49\textwidth}
         \centering
         \includegraphics[width=\textwidth]{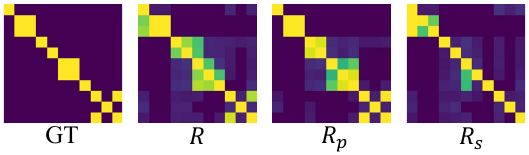}
         \caption{}
         \label{fig:R_a}
         \vspace{0.1cm}
     \end{subfigure}
     \begin{subfigure}[b]{0.49\textwidth}
         \centering
         \includegraphics[width=\textwidth]{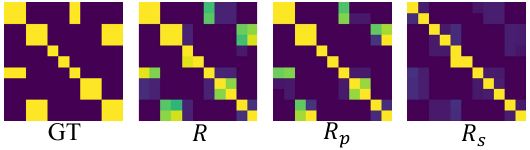}
         \caption{}
         \label{fig:R_b}
         \vspace{0.1cm}
     \end{subfigure}
     % \vspace{-0.3cm}
     }
    {
    \begin{subfigure}[b]{0.49\textwidth}
         \centering
         \includegraphics[width=\textwidth]{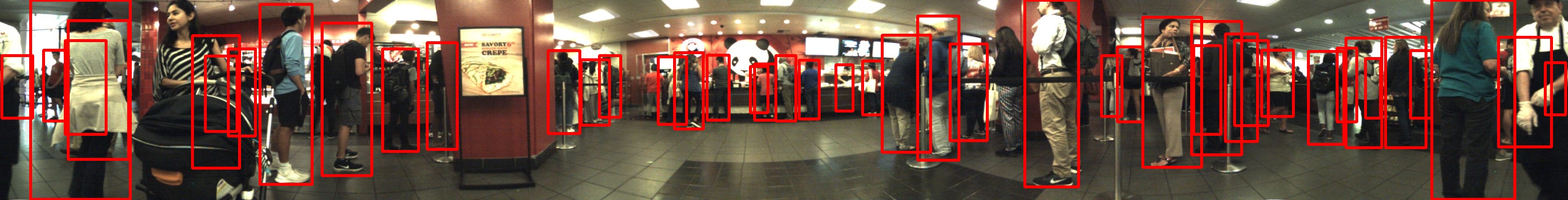}
         \vspace{-0.3cm}
     \end{subfigure}
     \begin{subfigure}[b]{0.49\textwidth}
         \centering
         \includegraphics[width=\textwidth]{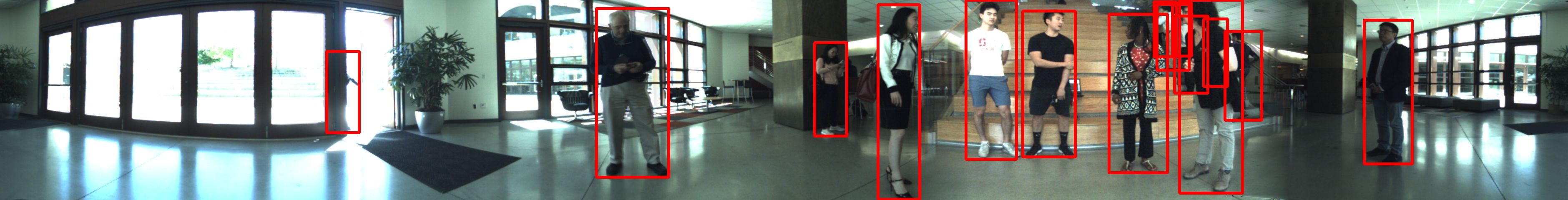}
         \vspace{-0.3cm}
     \end{subfigure}
     % \vspace{-0.3cm}
     }
    {
    \begin{subfigure}[b]{0.49\textwidth}
         \centering
         \includegraphics[width=\textwidth]{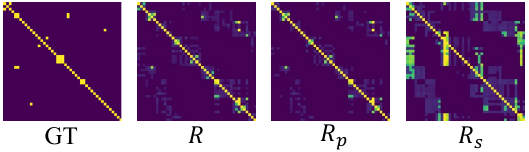}
         \caption{}
         \label{fig:R_c}
         \vspace{0.1cm}

     \end{subfigure}
     \begin{subfigure}[b]{0.49\textwidth}
         \centering
         \includegraphics[width=\textwidth]{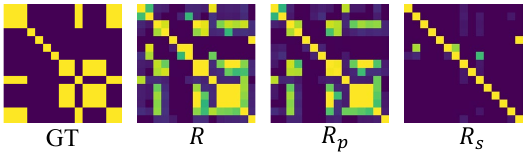}
         \caption{}
         \label{fig:R_d}
         \vspace{0.1cm}
     \end{subfigure}
     % \vspace{-0.3cm}
     }
    {
    \begin{subfigure}[b]{0.49\textwidth}
         \centering
         \includegraphics[width=\textwidth]{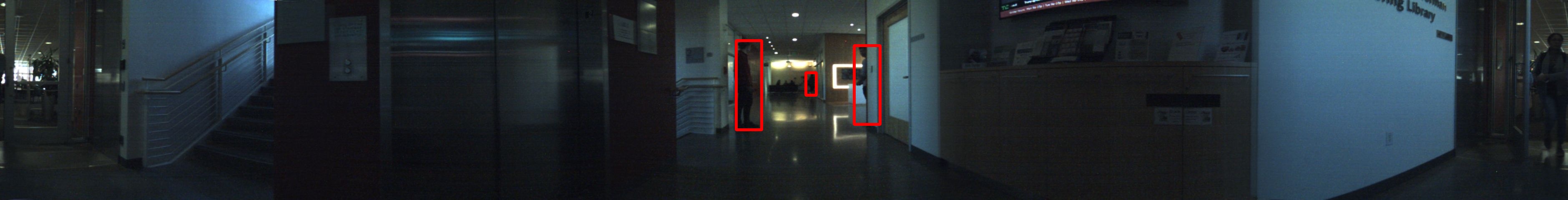}
         \vspace{-0.3cm}
     \end{subfigure}
     \begin{subfigure}[b]{0.49\textwidth}
         \centering
         \includegraphics[width=\textwidth]{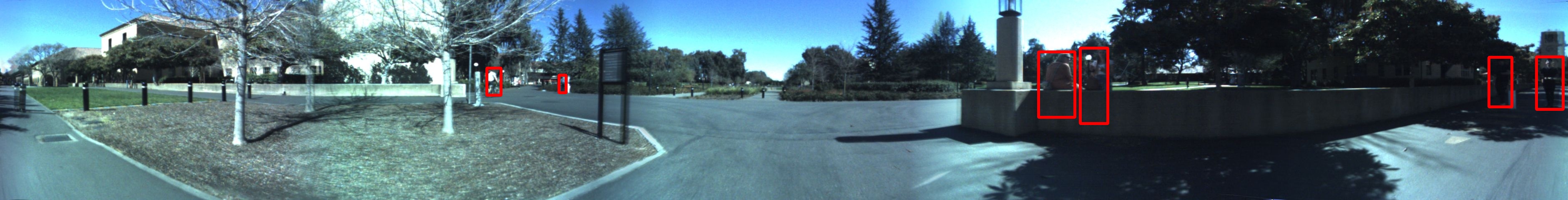}
         \vspace{-0.3cm}
     \end{subfigure}
     % \vspace{-0.3cm}
     }
    {
    \begin{subfigure}[b]{0.49\textwidth}
         \centering
         \includegraphics[width=\textwidth]{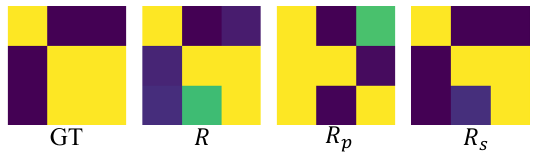}
         \caption{}
         \label{fig:R_e}
         \vspace{-0.3cm}

     \end{subfigure}
     \begin{subfigure}[b]{0.49\textwidth}
         \centering
         \includegraphics[width=\textwidth]{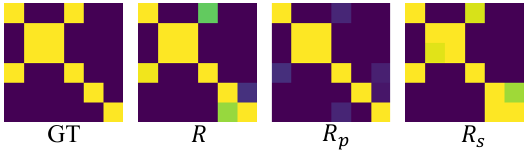}
         \caption{}
         \label{fig:R_f}
         \vspace{-0.3cm}
     \end{subfigure}
     % \vspace{-0.3cm}
     }
    \caption{Visualization of the ground-truth (GT) and predicted relation matrix $R$, the proximity relation matrix $R_p$, and the similarity matrix $R_s$. Best viewed zoomed in on screen.}
    \label{fig:relation_matrix}
    % \vspace{\belowfigcapmargin}
\end{figure}
%----------------------------------------------------------------------
\subsection{Loss functions}
We ablate the auxiliary loss $L_{aux}$ and the relation loss $L_{R}$ functions and summarize the results in Table~\ref{tab:loss}.
While $L_{aux}$ encourages the individual self-attention to learn individual action information, $L_{R}$ drives the visual similarity matrix $R_s$ to capture social relationships among individuals.
While solely using $L_R$ results in slight improvements in multi-granular activity recognition compared with the baseline, utilizing $\mathcal{L}_{aux}$ achieves performance improvements by 3.9\% in $\mathcal{F}_i$, 2.9\%p in $\mathcal{F}_p$, and 3.2\% in $\mathcal{F}_i$.
With cooperatively synergistic effects of $\mathcal{L}_{aux}$ and $\mathcal{L}_{R}$, \ournet\ achieves the best performance in both multi-granular activity recognition and social group detection.

\subsection{Balancing Hyper-parameters}
In Table~\ref{tab:lambda}, we summarize the results of experiments with varying balancing hyper-parameters of the individual action loss $\mathcal{L}_{idv}$, the relation loss $\mathcal{L}_{R}$, and the auxiliary loss $\mathcal{L}_{aux}$ functions.
When $\lambda_{idv}$ and $\lambda_{aux}$ are increased, we observe that a slight improvement in IAR and GBAR, but a significant performance decrease in SGAR.
Conversely, when $\lambda_{R}$ is increased, the overall performance $\mathcal{F}_a$ is decreased by 1.6\%.
When the proportions of $\lambda_{idv}$, $\lambda_{R}$, and $\lambda_{aux}$ are equal, \ournet\ achieves the best performance in social group detection performance and overall multi-granular activity recognition ($\mathcal{F}_a$).

\section{Additional Visualization}
\subsection{Relation Matrix}
Figure~\ref{fig:relation_matrix} shows more visual comparisons between the ground truth and predicted social relation matrix $R$ with the proximity relation matrix $R_p$ and the similarity matrix $R_s$.
Those matrices have 1 for individuals belonging to the same social group and 0 for otherwise.
We note that $R_p$ closely corresponds to the ground-truth social relation compared to $R_s$ (see Fig.~\ref{fig:R_a}, \ref{fig:R_b}, \ref{fig:R_c}, and \ref{fig:R_d}).
In contrast, in panoramic scenes with relatively larger bounding boxes and fewer people, we observe that $R_s$ is effective (see Fig.~\ref{fig:R_e} and~\ref{fig:R_f}).

\subsection{Prediction Results}
\begin{figure}[t]     
    \begin{subfigure}[t]{0.95\textwidth}
        \centering
        \includegraphics[width=\textwidth]{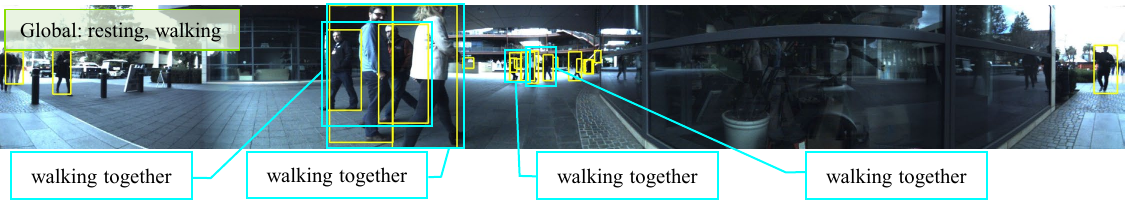}
        \caption{GT}
    \end{subfigure}
    
    \begin{subfigure}[t]{0.95\textwidth}
        \includegraphics[width=\textwidth]{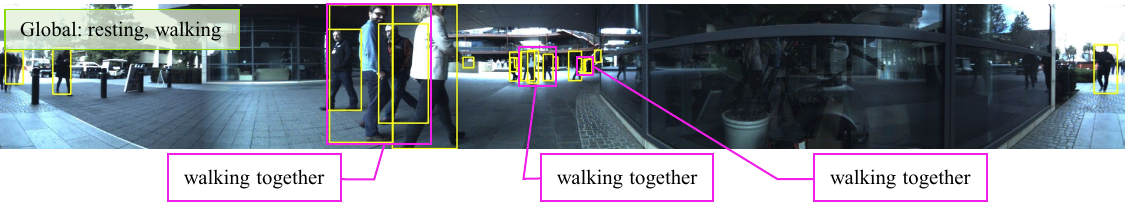}
        \caption{without the proximity relation $R_p$}
    \end{subfigure}
    
    \begin{subfigure}[t]{0.95\textwidth}
        \includegraphics[width=\textwidth]{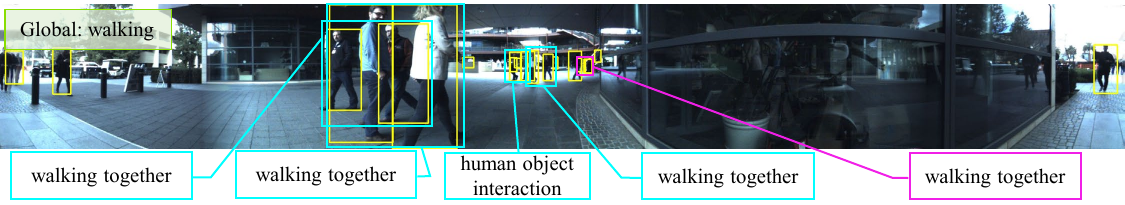}
        \caption{Ours}
    \end{subfigure}
    \vspace{-0.3cm}
    \caption{Visual comparisons of the social group activity detection and global activity recognition between of (a) ground-truth, (b) \ournet\ without the social proximity relation $R_p$, and (c) \ournet. Misclassified social group detections are indicated in \textcolor{magenta}{magenta}, while ground-truth and correctly predicted bounding boxes are in \textcolor{cyan}{cyan}.}
    \label{fig:result1}
\end{figure}

\begin{figure}[t]     
    \begin{subfigure}[t]{0.95\textwidth}
        \centering
        \includegraphics[width=\textwidth]{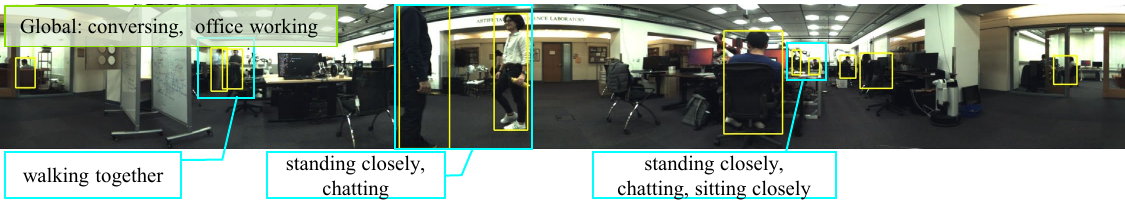}
        \caption{GT}
    \end{subfigure}
    
    \begin{subfigure}[t]{0.95\textwidth}
        \includegraphics[width=\textwidth]{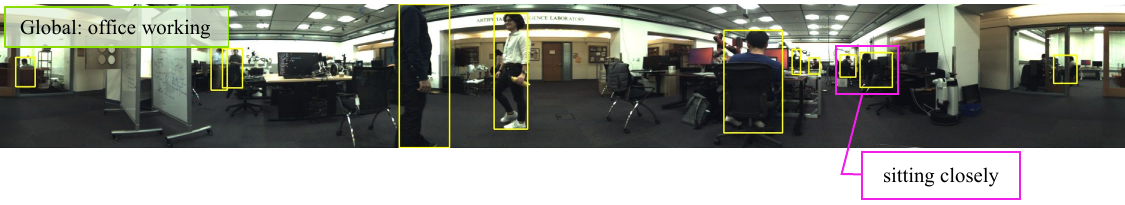}
        \caption{without the proximity relation $R_p$}
    \end{subfigure}
    
    \begin{subfigure}[t]{0.95\textwidth}
        \includegraphics[width=\textwidth]{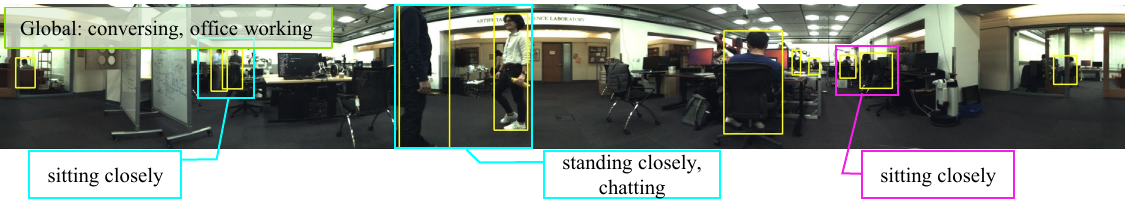}
        \caption{Ours}
    \end{subfigure}
    \vspace{-0.3cm}
    \caption{Visual comparisons of the social group activity detection and global activity recognition between of (a) ground-truth, (b) \ournet\ without the social proximity relation $R_p$, and (c) \ournet. Misclassified social group detections are indicated in \textcolor{magenta}{magenta}, while ground-truth and correctly predicted bounding boxes are in \textcolor{cyan}{cyan}.}
    \label{fig:result2}
\end{figure}

\begin{figure}[t]     
    \begin{subfigure}[t]{0.95\textwidth}
        \centering
        \includegraphics[width=\textwidth]{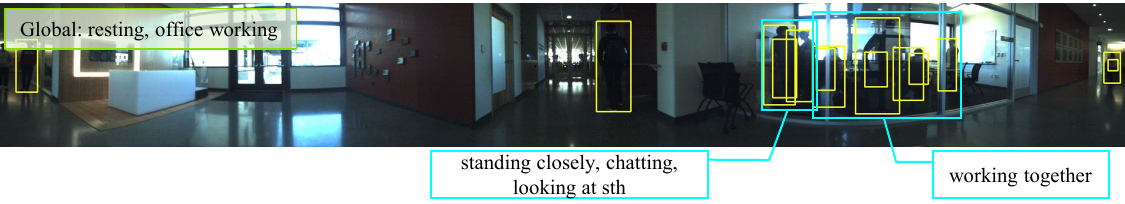}
        \caption{GT}
    \end{subfigure}
    
    \begin{subfigure}[t]{0.95\textwidth}
        \includegraphics[width=\textwidth]{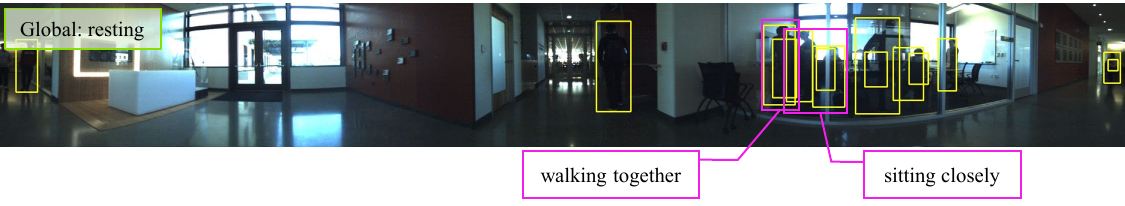}
        \caption{without the proximity relation $R_p$}
    \end{subfigure}
    
    \begin{subfigure}[t]{0.95\textwidth}
        \includegraphics[width=\textwidth]{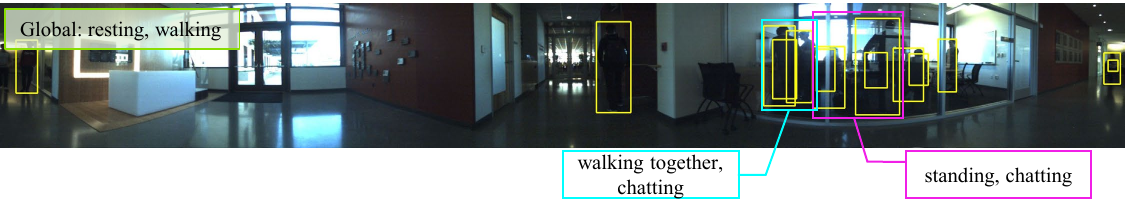}
        \caption{Ours}
    \end{subfigure}
    \vspace{-0.3cm}
    \caption{Visual comparisons of the social group activity detection and global activity recognition between of (a) ground-truth, (b) \ournet\ without the social proximity relation $R_p$, and (c) \ournet. Misclassified social group detections are indicated in \textcolor{magenta}{magenta}, while ground-truth and correctly predicted bounding boxes are in \textcolor{cyan}{cyan}.}
    \label{fig:result3}
\end{figure}

\begin{figure}[t]     
    \begin{subfigure}[t]{0.95\textwidth}
        \centering
        \includegraphics[width=\textwidth]{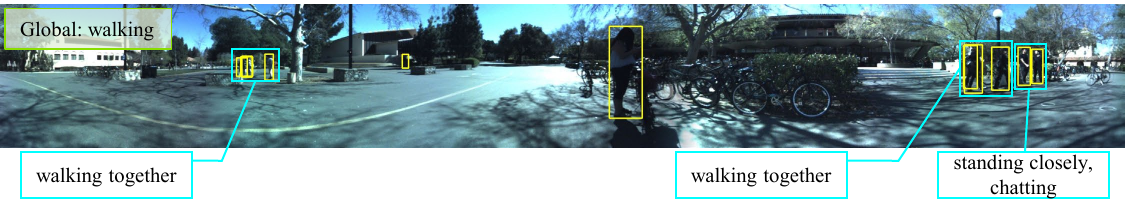}
        \caption{GT}
    \end{subfigure}
    
    \begin{subfigure}[t]{0.95\textwidth}
        \includegraphics[width=\textwidth]{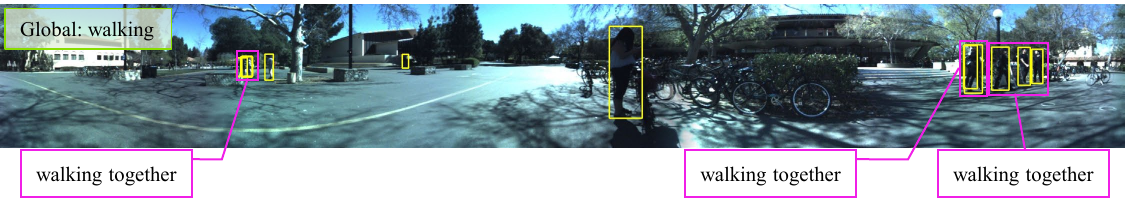}
        \caption{without the proximity relation $R_p$}
    \end{subfigure}
    
    \begin{subfigure}[t]{0.95\textwidth}
        \includegraphics[width=\textwidth]{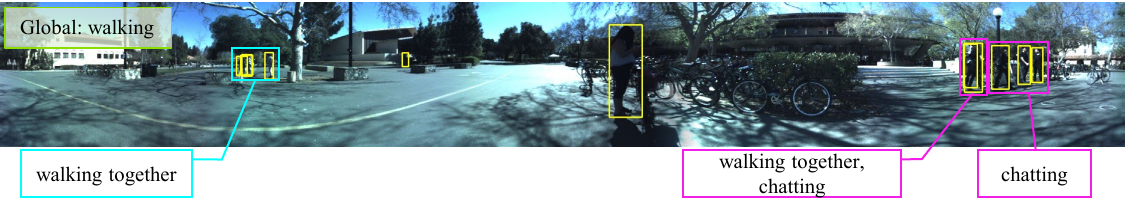}
        \caption{Ours}
    \end{subfigure}
    \vspace{-0.3cm}
    \caption{Visual comparisons of the social group activity detection and global activity recognition between of (a) ground-truth, (b) \ournet\ without the social proximity relation $R_p$, and (c) \ournet. Misclassified social group detections are indicated in \textcolor{magenta}{magenta}, while ground-truth and correctly predicted bounding boxes are in \textcolor{cyan}{cyan}.}
    \label{fig:result4}
\end{figure}

Figure~\ref{fig:result1}, \ref{fig:result2}, \ref{fig:result3}, and \ref{fig:result4} shows the visual comparisons between the ground-truth and \ournet\ with and without the social proximity relation $R_p$ on JRDB-PAR dataset~\cite{par_eccv}.
We note that the absence of utilizing $R_p$ leads to inaccurate or missed detections of social groups.

\section{Limitation and Future Work.} There are still unresolved problems.
In Table \textcolor{red}{6} in the manuscript, we observed that the performance of social group activity detection is enhanced when using the ground-truth number of social groups.
To address this, it is necessary to develop strategies to adjust to varying group densities and complexities.
Moreover, real-world datasets, such as JRDB-PAR~\cite{par_eccv}, exhibit significant biases in their class distributions.
Overcoming these biases is crucial for improving the robustness and generalization of the proposed method.
We leave this intriguing challenge to future work.

%======================================================================
%----------------------------------------------------------------------
%======================================================================
%----------------------------------------------------------------------

%======================================================================
%----------------------------------------------------------------------
%======================================================================
%----------------------------------------------------------------------

% \input{chapter_supp/01_introduction}

% \clearpage

\end{document}